\documentclass[10pt,twocolumn,letterpaper]{article}

\usepackage[pagenumbers]{cvpr} 

\usepackage{graphicx}
\usepackage{latexsym}

\usepackage{times}
\usepackage{soul}
\usepackage{url}
\usepackage[utf8]{inputenc}
\usepackage{algorithm}
\usepackage{algorithmic}
\usepackage[switch]{lineno}

\usepackage{amsmath,amsfonts,amsthm, amssymb, enumerate}
\usepackage{lipsum}
\usepackage{graphicx}
\usepackage[bottom]{footmisc}
\usepackage[table, dvipsnames]{xcolor}
\usepackage{array,multirow}
\usepackage{enumitem}

\usepackage{microtype}
\usepackage{booktabs}
\usepackage{pifont}

\usepackage{times}
\usepackage{epsfig}
\usepackage{enumitem}
\usepackage{setspace}
\usepackage{wrapfig}
\usepackage{extarrows}
\usepackage{makecell}
\usepackage{bbm}  
\usepackage{textcomp}
\usepackage{xcolor}
\usepackage{caption}

\newcommand{\argmax}{\operatornamewithlimits{argmax}}

\newcommand{\minmax}{\operatornamewithlimits{mnmx}}
\newcommand{\average}{\operatornamewithlimits{avg}}

\newcolumntype{M}[1]{>{\centering\arraybackslash}m{#1}}
\newcolumntype{N}{@{}m{0pt}@{}}

\definecolor{maroon}{cmyk}{0,0.87,0.68,0.32}

\definecolor{cvprblue}{rgb}{0.21,0.49,0.74}
\usepackage[pagebackref,breaklinks,colorlinks,citecolor=cvprblue]{hyperref}

\def\paperID{5967} 
\def\confName{CVPR}
\def\confYear{2024}

\title{RecurSeed and EdgePredictMix: Pseudo-Label Refinement Learning for Weakly Supervised Semantic Segmentation across Single- and Multi-Stage Frameworks}

\author{First Author\\
Institution1\\
Institution1 address\\
{\tt\small firstauthor@i1.org}
\and
Second Author\\
Institution2\\
First line of institution2 address\\
{\tt\small secondauthor@i2.org}
}

\author{
    Sanghyun Jo$^{1}$ \qquad In-Jae Yu$^{2}$ \qquad Kyungsu Kim$^{3}$\thanks{Correspondence to}\\ 
    {$^{1}$OGQ, Seoul, Korea} \qquad {$^{2}$Samsung Electronics, Suwon, Korea}\\
    {$^{3}$Department of Data Convergence and Future Medicine, Sungkyunkwan University, Seoul, Korea}\\ 
  \texttt{\{shjo.april, ijyu.phd, kskim.doc\}@gmail.com}\\
}

\begin{document}
\maketitle

\begin{abstract}
Although weakly supervised semantic segmentation using only image-level labels (WSSS-IL) is potentially useful, its low performance and implementation complexity still limit its application. The main causes are (a) non-detection and (b) false-detection phenomena: (a) The class activation maps refined from existing WSSS-IL methods still only represent partial regions for large-scale objects, and (b) for small-scale objects, over-activation causes them to deviate from the object edges. We propose RecurSeed, which alternately reduces non- and false detections through recursive iterations, thereby implicitly finding an optimal junction that minimizes both errors. We also propose a novel data augmentation (DA) approach called EdgePredictMix, which further expresses an object's edge by utilizing the probability difference information between adjacent pixels in combining the segmentation results, thereby compensating for the shortcomings when applying the existing DA methods to WSSS. We achieved new state-of-the-art performances on both the PASCAL VOC 2012 and MS COCO 2014 benchmarks (VOC \emph{val}: $74.4\%$, COCO \emph{val}: $46.4\%$). The code is available at \url{https://github.com/shjo-april/RecurSeed_and_EdgePredictMix}.
\end{abstract}

\vspace{-0.5cm}

\section{Introduction}
\label{section:intro}


{Semantic segmentation is pivotal in computer vision image analysis. However, training these models demands pixel-wise annotations, taking about 20 seconds for image-level and 239 seconds for pixel-level annotations per image. Recent methods \cite{ru2023token,kweon2023weakly} have developed weakly supervised semantic segmentation (WSSS) techniques leveraging weaker labels to lower the cost of mask labeling, like image-level class labels \cite{ahn2018learning} and points \cite{bearman2016s}. Among these, we focus on WSSS using image-level class labels (WSSS-IL) as the most economical supervision.}

\begin{figure}[t]
  \centering
  \includegraphics[width=0.95\linewidth, height=0.325\linewidth]{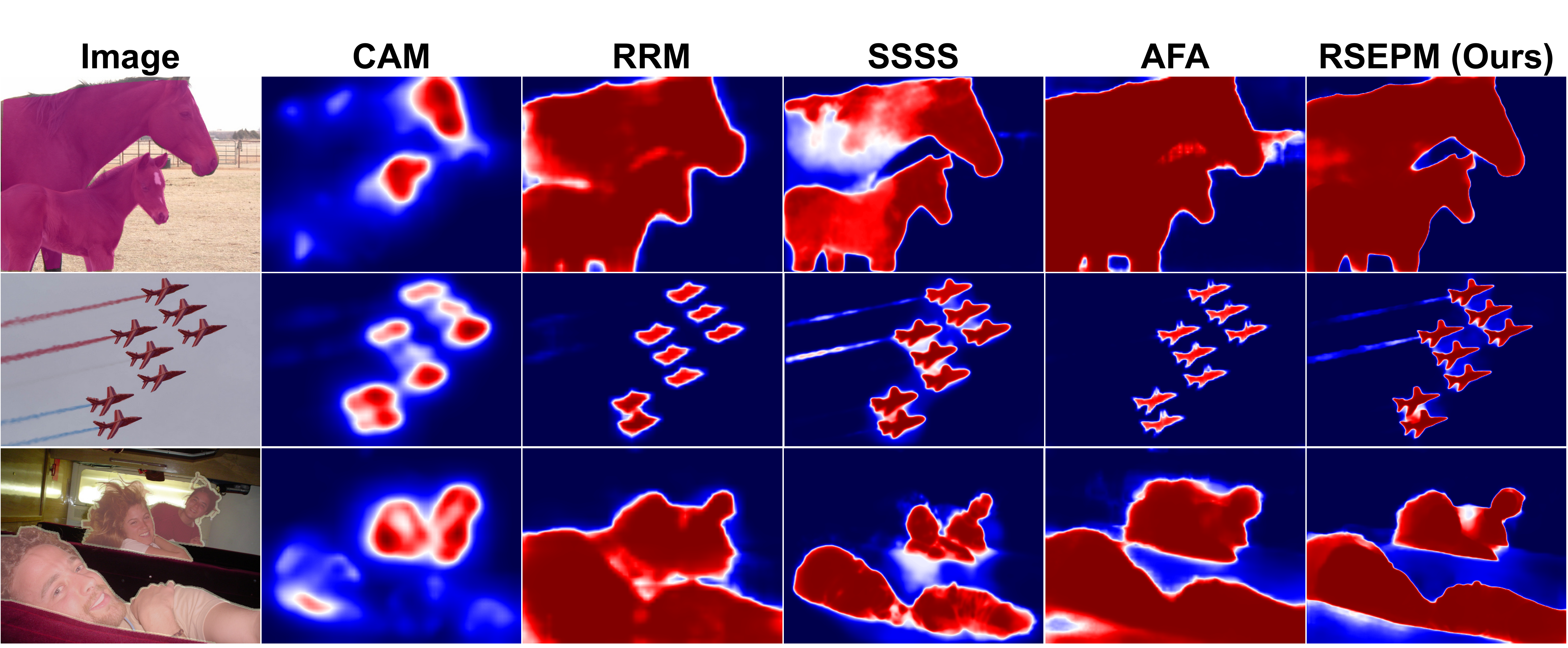}
  \vspace{-0.3cm}
  \caption{
      Comparison of localization maps produced by using conventional CAMs, existing single-stage learning frameworks, and ours, with PASCAL VOC 2012 training set. Our results show foreground and background regions more accurately than other localization maps, thanks to \textit{the recursive learning process} (RS).
  }
  \label{fig:intro}
  \vspace{-0.5cm}
\end{figure}


{Most WSSS-IL approaches utilize a class activation map (CAM) \cite{zhou2016learning}. However, CAM highlights the object's most discriminative regions, causing many false negatives (FNs). To mitigate CAM's limitations, recent state-of-the-art methods \cite{zhou2022regional,chen2022self,cheng2023out} adopt a multi-stage learning framework (MLF). This MLF encompasses stages tailored for specific tasks, rendering the training process intricate. Single-stage learning frameworks (SLFs) address these MLF complexities by employing a unified model. Nevertheless, current SLFs \cite{roy2017combining,zhang2020reliability} lag in performance compared to MLFs.}


{Meanwhile, post-processing modules for CAMs, such as Self-Correlation map Generation (SCG) \cite{pan2021unveiling} and Pixel-Adaptive Mask Refinement (PAMR) \cite{araslanov2020single}, leverage feature and image correlations, respectively. While SCG enhances spatial details for larger objects, it raises false positives (FPs) for medium and small objects. In contrast, PAMR effectively reduces FPs by refining the CAM's boundaries, but when used alone, it can increase FNs due to a lack of dispersion in initial CAMs. These conflicting FN and FP issues in WSSS-IL are termed FPN-WSSS-IL.}


{On the other hand, data augmentation (DA) is a straightforward technique that boosts data efficacy. Existing DA strategies, like ClassMix \cite{olsson2021classmix}, enrich datasets by blending images using model-predicted masks. However, in the WSSS-IL context, implementing such mixing augmentations without refining these predicted masks escalates ambiguity stemming from the partial object boundaries present in the masks. We denote this challenge as IBDA-WSSS-IL.}

\begin{table}
  \caption{  
    Comparison with {RS and its related works to highlight our novelty}: SEAM \protect\cite{wang2020self}, RRM \protect\cite{zhang2020reliability}, AFA \protect\cite{ru2022learning}.
  } 
  \vspace{-0.2cm}
  \centering
  \begin{scriptsize} 
  \begin{tabular}{p{0.275\textwidth} || p{0.02\textwidth} p{0.02\textwidth} p{0.02\textwidth} p{0.025\textwidth}}
    \toprule
    Properties & SEAM & RRM & AFA & \textbf{RS (Ours)} \\  
    \hline 
    Employ recursion principle for improving CAM & \checkmark & \checkmark & \checkmark & \checkmark \\
    Reduce FP by using mask refinement &  & \checkmark & \checkmark & \checkmark \\
    Reduce FN by using the first-order correlation & \checkmark &  & \checkmark & \checkmark \\
    Reduce FN by using the second-order correlation &  &  &  & \checkmark \\
    \bottomrule
  \end{tabular}
  \label{tab:novelty_rs}
  \end{scriptsize}
  \vspace{-0.3cm}
\end{table}



{To address both FPN-WSSS-IL and IBDA-WSSS-IL challenges, we introduce RecurSeed (RS) and EdgePredictMix (EPM). RS recursively refines the initial CAM by leveraging its high-order correlation (\emph{e.g.}, SCG) along with mask refinements (\emph{e.g.}, PAMR) during training, thereby diminishing both FP and FN in CAM, as illustrated in Fig. \ref{fig:intro}. EPM, the pioneering DA solution for IBDA-WSSS-IL, enhances uncertain pixels in mixed results by restoring edge data from the model's predictions. Our primary contributions can be summarized as follows.}


\begin{itemize}
     \item {To mitigate FPN-WSSS-IL, we \textit{newly apply a recursion} (RS) to the integration of SCG and PAMR beyond their original post-processing roles. Unlike a simple combination for post-processing, our RS facilitates the network's convergence towards a mask that optimally reduces both FN and FP, yielding a performance boost over 11\% compared to non-recursive methods. The efficacy of RS stems from recursively transferring the knowledge of spatial information (PAMR) refined by the second-order correlation (SCG) to the network. By leveraging the high-order correlation, our RS surpasses traditional recursion methods (\emph{i.e.}, using the first-order correlation), improving the performance by 7.4\% in mIoU (see Tab. \ref{tab:ordering}).}
     \item {To address IBDA-WSSS-IL, we propose a novel edge refinement (EP) optimized for mixing augmentation (EPM). To maximize the mixing effect, EPM refines predicted masks \textit{by jointly harnessing absolute and relative per-pixel probability values}, consistently boosting the performance of various WSSS methods in Tab. \ref{tab:epm}.}
     \item {Our SLF achieves the highest performance with an mIoU of 70.6\% on the VOC 2012 test set, even without utilizing advanced backbones (\emph{e.g.}, transformers). This result demonstrates that our SLF can deliver superior performance in WSSS-IL without intricate learning configurations like MLF. When our SLF is extended straightforwardly into MLF by applying a random walk \cite{ahn2018learning}, our method exceeds recent MLFs \cite{xu2022multi,cheng2023out} by a 1.7\% and surpasses recent methods using additional supervision, like saliency \cite{jiang2022l2g}, on all benchmarks (see Tab. \ref{tab:comp_voc_coco}).}
\end{itemize}

\section{Related work}
\label{section:related}

\subsection{Weakly supervised semantic segmentation}

Most studies \cite{kolesnikov2016seed,lee2019ficklenet,sun2020mining,lee2021anti,wu2021embedded,jo2021puzzle,lee2021reducing,jiang2022l2g,du2022weakly,zhou2022regional,lee2022threshold,xu2022multi,kweon2023weakly,rong2023boundary,cheng2023out} have focused on improving the initial CAM quality, adopting MLF that involves producing an initial CAM, generating pseudo masks, and training the segmentation model. Independent research like $C^2$AM \cite{xie2022c2am}, ADEHE \cite{liu2022adaptive}, and SANCE \cite{li2022towards} aimed to rectify inaccuracies in pseudo masks. Meanwhile, W-OoD \cite{lee2022weakly} and CLIMS \cite{xie2022clims} employed additional supervision, such as saliency and manually collected dataset, to address CAM's limitations. All studies above did not consider a recursion process.


\vspace{-0.4cm}

\paragraph{Novelty of RS.} {While several methods \cite{wang2020self,zhang2021complementary,chen2022self,zhang2020reliability,araslanov2020single,ru2022learning,ru2023token} have integrated the recursive process akin to our RecurSeed (RS), they have yet to leverage high-order correlations, causing an inadequate reduction in FN. In Tab. \ref{tab:novelty_rs}, our RS stands out in key characteristics. Specifically, approaches like SEAM \cite{wang2020self}, CPN \cite{zhang2021complementary}, and SIPE \cite{chen2022self} utilize first-order correlations without mask refinement, increasing FP with limited FN reductions for CAMs. Meanwhile, RRM \cite{zhang2020reliability} and SSSS \cite{araslanov2020single} refine the typical CAM in the training loop without any feature correlation, leaving the CAM's FN unaddressed. Employing a transformer architecture, AFA \cite{ru2022learning} developed an SLF using first-order correlation and mask refinement. Despite leveraging the inherent strengths of transformers \cite{ru2023token}, the first-order correlation leads to only a slight FN reduction in their approach. Our study also belongs to an SLF but surprisingly mitigates FP and FN \textit{by jointly applying the high-order correlation of CAM and its subsequent refinement in the recursion principle for the first time}.}

\begin{table}[t] 
  \caption{  
    Comparison with EPM and its related DA studies to highlight our novelty: SG \protect\cite{park2022saliency}, CDA \protect\cite{su2021context}, ClassMix \protect\cite{olsson2021classmix}.
  } 
  \vspace{-0.2cm}
  \centering
  \begin{scriptsize} 
  \begin{tabular}{p{0.23\textwidth} || c c c p{0.025\textwidth}}
    \toprule
    Properties & SG & CDA & ClassMix & \textbf{EPM (Ours)} \\  
    \hline 
    Use predicted mask in mix &  \checkmark &  \checkmark &  \checkmark &  \checkmark \\
    Use mixed mask in learning  &  &  &  \checkmark &  \checkmark \\
    Consider mix in WSSS &  &  \checkmark &  &  \checkmark \\
    Refine predicted mask using edge &  &  &  &  \checkmark \\
    \bottomrule
  \end{tabular}
  \label{tab:seed}
  \end{scriptsize}
  \vspace{-0.6cm}
\end{table}

\subsection{Data augmentation}
CutMix \cite{yun2019cutmix} crops a random region of an image and pastes it on another. Saliency Grafting (SG) \cite{park2022saliency} adjusts mixed class labels based on occlusion degrees derived from the CAM's saliency map. CDA \cite{su2021context} is the first DA method for improving WSSS performance. ClassMix \cite{olsson2021classmix} merges unlabeled images with decoder outputs by training partial pixel-wise annotations.


\vspace{-0.4cm}

\paragraph{Novelty of EPM.} Previous works have employed affine transformation \cite{wang2020self}, cropping \cite{jiang2022l2g}, and mixing \cite{su2021context}. Our EdgePredictMix (EPM) is the sole method considering edge refinement to amplify the mixing effect. Specifically, our edge refinement (EP) technique enhances ambiguous pixels using edge data from uncertain regions in predicted masks. Tab. \ref{tab:seed} compares our EPM with other DA methods regarding critical properties.

\section{Method}

\begin{figure*}[t]%
    \centering%
    \includegraphics[width=1.00\linewidth]{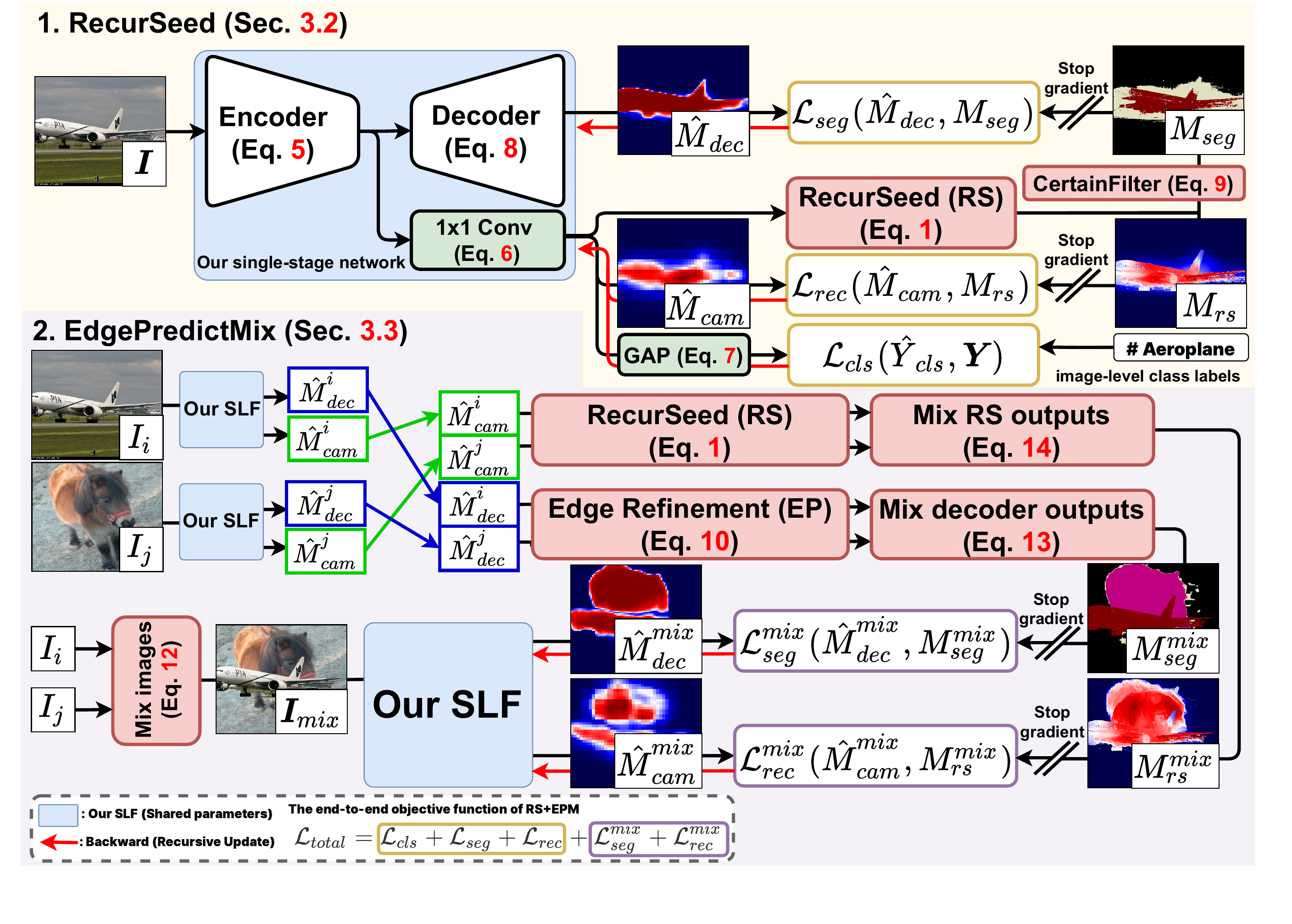}
    \vspace{-0.5cm}
    \caption{ 
        Overview of our single-stage learning framework (SLF) with the proposed RS and EPM. Based on extracted features from an encoder (\emph{e.g.}, ResNet-50 \cite{he2016deep}), the classifier and decoder (\emph{e.g.}, DeepLabv3+ \cite{chen2018encoder}) produce CAMs $\hat{M}_{cam}$ in \eqref{cam} and per-pixel segmentation outputs $\hat{M}_{dec}$ in \eqref{dec_output}, respectively. RS provides refined CAMs $M_{rs}$ in \eqref{mrs} and pseudo masks $M_{seg}$ in \eqref{certainfilter} by utilizing the high-order (\emph{e.g.}, SCG) and image (\emph{e.g.}, PAMR) correlations. EPM synthesizes two images $\boldsymbol{I}_{mix}$, RS outputs $M^{mix}_{rs}$, and decoder outputs $M^{mix}_{seg}$ refined by EP in \eqref{ep_alg} to improve WSSS performance further. Our network also produces mixed predictions ($\hat{M}^{mix}_{cam}$ and $\hat{M}^{mix}_{seg}$) from a mixed image. Finally, our encoder and decoder recursively update with them by using RS-related ($\mathcal{L}_{cls}$ \eqref{recurseed_label}, $\mathcal{L}_{seg}$ \eqref{recurseed_seg}, and $\mathcal{L}_{rec}$ \eqref{enc_rec_loss}) and EPM-related losses  ($\mathcal{L}^{mix}_{seg}$ \eqref{mix_seg} and $\mathcal{L}^{mix}_{rec}$ \eqref{mix_rec}).
    }
    \label{fig:overview}
    \vspace{-0.2cm}
\end{figure*}

\subsection{Overview} 
Our proposed method for the single-stage learning framework encompasses parallel classification and segmentation branches. We introduce two components tailored to WSSS: (i) RecurSeed (RS) and (ii) EdgePredictMix (EPM). RS iteratively updates the initial CAM, leveraging high-order and image correlations, enlarging previously undetected foregrounds. Our edge refinement (EP) removes uncertain pixels in pseudo masks using the edge details from model predictions. Subsequently, EPM merges pairs of images and EP-refined pseudo masks, facilitating sample diversity and enhancing WSSS performance. The comprehensive framework is depicted in Fig. \ref{fig:overview}.

\subsection{RecurSeed}
\label{section:recurseed}
\begin{figure}[t]%
    \centering%
    \includegraphics[width=1.0\linewidth, height=0.8\linewidth]{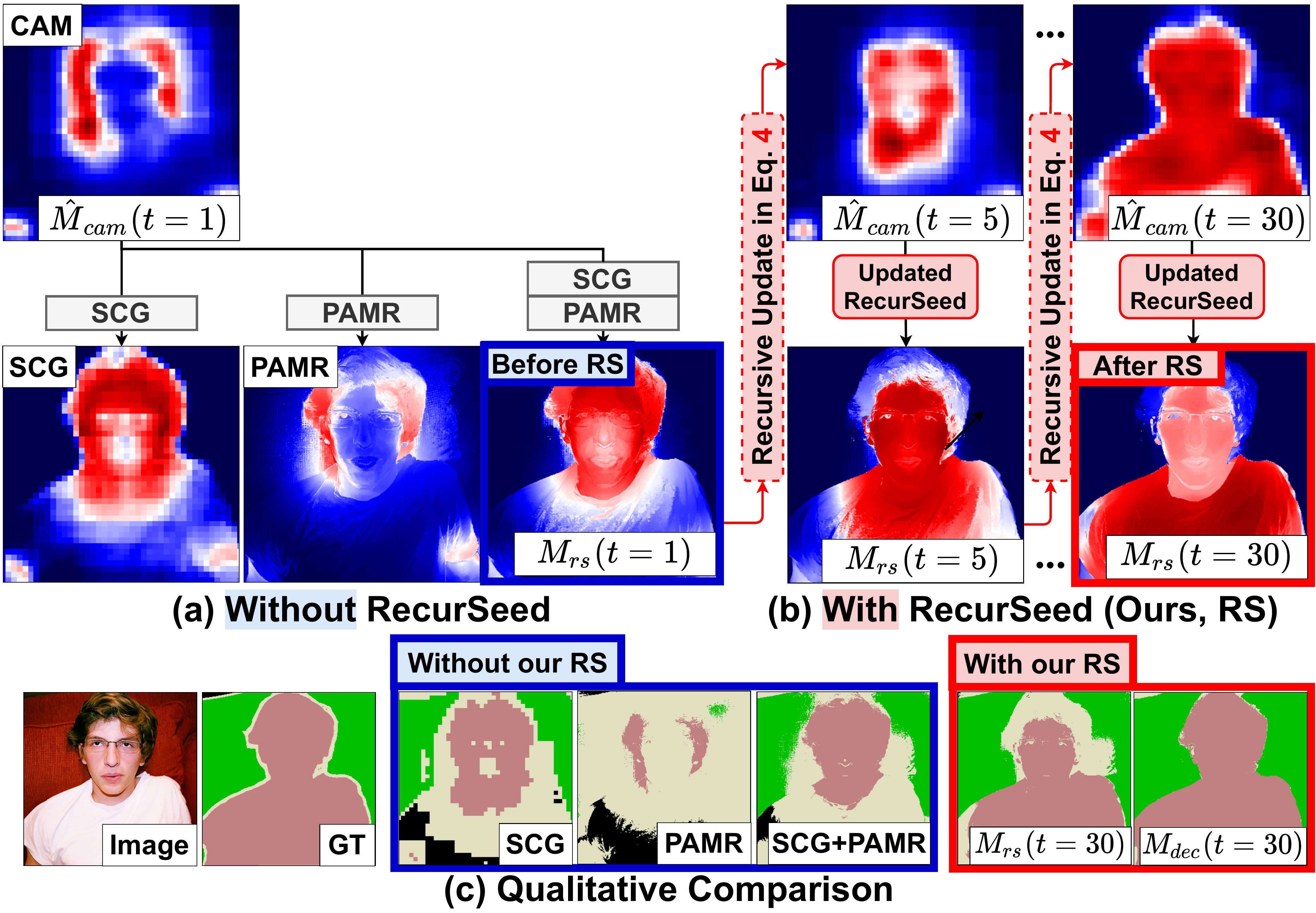}
    \vspace{-0.7cm}
    \caption{
        (a) Without our RS, the simple post-processing schemes for CAM show each output. (b) With our RS, our method iteratively improves the CAM, thereby taking advantage of both the strengths of SCG (\emph{i.e.}, reducing FN) and PAMR (\emph{i.e.}, reducing FP). (c) Quality of pseudo masks produced from CertainFilter (CF) in \eqref{certainfilter}. Simple post-processing results (\emph{e.g.}, SCG) fail to detect integral objects precisely. By contrast, our method delineates target objects by updating the initial CAM \textit{recursively}.
    }
    \label{fig:rs}
    \vspace{-0.3cm}
\end{figure}

{Our application of recursive learning to baselines that utilize high-order (\emph{e.g.}, SCG) and image correlations (\emph{e.g.}, PAMR) adeptly mitigates FPN-WSSS-IL by imparting the semantic knowledge of refined CAMs onto the network. Without our RS, juxtaposing SCG and PAMR yields only a modest enhancement in CAM quality, as evidenced in Fig. \ref{fig:rs}(a). This approach still results in a high FN, attributed to the suboptimal initial CAMs. Our RS resolves the limitations from the direct amalgamation of SCG and PAMR by putting them firstly to the learning process, as illustrated in Fig. \ref{fig:rs}(b).}

In this section, we introduce RS $M_{rs}$ as follows: For the $t$-th epoch, where $t \in \{0:T\}$,
\begin{align}\label{mrs}
&M_{rs}(t) = PAMR(SCG(M_{cam}(t));\mathcal{W}), 
\end{align} 
where $PAMR$ and $SCG$ refer to their processes respectively (detailed in Appendix A), $M_{cam}(0)$ is the initial CAM, $M_{rs}(t)$ denotes {its refined result through RS}, and {$\mathcal{W}=SCG(M_{cam}(t))$ is set as the SCG's activation region}. Thus, $M_{rs}(0)$ expressed in \eqref{mrs} is simply the result of sequentially applying SCG and PAMR to the CAM's result. In particular, our contribution is that we add the recursive learning as in \eqref{enc_rec_loss} so that CAM can reconstruct this result (RS); thanks to RS, the CAM results gradually improve as learning progressed, beyond this simple integration $M_{rs}(0)$. 
Intuitively, we recursively improve the result of $M_{cam}(0)$ by training the network such that the CAM updated at the next step $M_{cam}(1)$ becomes the result $M_{rs}(0)$, aiming at $M_{rs}(t) \approx M_{cam}(t+1)$ for every epoch $t \in \{0:T\}$. As shown in Figs \ref{fig:rs}(b) and (c), our RS gradually updates the initial seed (\emph{i.e.}, CAMs) to remedy the shortcomings of the simple combination of SCG and PAMR in \eqref{mrs}.

To achieve the objective of $M_{rs}(t) \approx M_{cam}(t+1)$, we minimize the reconstruction loss in \eqref{recurseed_seg} (\emph{i.e.}, in decoder domain) and \eqref{enc_rec_loss} (\emph{i.e.}, in encoder domain) along with the classification loss in \eqref{recurseed_label}, and the total net parameter $\theta_{t}$ is updated at step $t$ as follows:
\begin{small}
\begin{align}\label{recurseed_label}
\theta_{t}   =  \theta_{t-1}  &- \eta \frac{\partial}{\partial{\theta}}\mathbb{E}_{\boldsymbol{I}}\big[\mathcal{L}_{cls}\big(\hat{Y}_{cls}(\tau), \boldsymbol{Y}; \theta \big)\big]\Big|_{\theta = \theta_{{\tau}},\tau=t-1} \\\label{recurseed_seg}
&- \eta \frac{\partial}{\partial{\theta}}\mathbb{E}_{\boldsymbol{I}}\big[ \mathcal{L}_{seg} \big(\hat{M}_{dec}(\tau),M_{seg}(\tau); \theta \big)\big]\Big|_{\theta = \theta_{{\tau}},\tau=t-1}\\\label{enc_rec_loss}
&- \eta \frac{\partial}{\partial{\theta}}\mathbb{E}_{\boldsymbol{I}}\big[ \mathcal{L}_{rec} \big(\hat{M}_{cam}(\tau),M_{rs}(\tau); \theta \big)\big]\Big|_{\theta = \theta_{{\tau}},\tau=t-1},
\end{align}
\end{small} 
where $\boldsymbol{I}=\{I^{1},I^{2},...,I^{B}\}$ denotes a mini-batch of size $B$, $\boldsymbol{Y}$ denotes their truth class labels,  
\begin{small}
\begin{align}\label{enc_output}
&f_{enc}(t) = E_{\theta^{enc}_{t}}(\boldsymbol{I}), \\\label{cam}
&\hat{M}_{cam}(t) = A_{\theta^{cls}_{t}}(f_{enc}(t)), \\\label{label_est}
&\hat{Y}_{cls}(t) = \sigma(GAP(\hat{M}_{cam}(t))), \\\label{dec_output}
&\hat{M}_{dec}(t) = D_{\theta^{dec}_{t}}(f_{enc}(t)),
\end{align}
\end{small}
and 
\begin{small}
\begin{align}\label{certainfilter}
 &M_{seg}(t) =  CF(M_{rs}(t)) \\\nonumber 
 &=  
 \begin{cases}
    \underset{c \in \mathcal{C}}{\argmax} \big(M^c_{rs}(t)[:,i,j]\big) & \text{if } \underset{c \in \mathcal{C}}{\max} \big(M^c_{rs}(t)[:,i,j]\big) > \delta_{fg},  \\
    0 & \text{if } \underset{c \in \mathcal{C}}{\max} \big(M^c_{rs}(t)[:,i,j]\big) < \delta_{bg}, \\ 
    -1  & \text{otherwise}.
\end{cases}
\end{align}
\end{small}
where 0 and -1 represent the background and ignored classes, respectively. As shown in Fig. \ref{fig:overview}, our network consists of an encoder $E(\cdot)$ and a decoder $D(\cdot)$ with outputs of $f_{enc}(t)$ in \eqref{enc_output} and $\hat{M}_{dec}(t)$ in \eqref{dec_output}, respectively. The refined CAM $\hat{M}_{cam}(t)$ is then obtained as $\eqref{cam}$ by adding layer $A(\cdot)$, scaling the number of channels to the number of classes. We apply global average pooling (GAP) and sigmoid $\sigma$ to estimate the class labels as $\hat{Y}_{cls}(t)$ in \eqref{label_est}. Following the common practice, we employ the multi-label soft margin loss for the classification loss $L_{cls}$, the per-pixel cross-entropy loss for the segmentation loss $L_{seg}$, and the L1 loss for the reconstruction loss $L_{rec}$.

Our network narrows the gap between the decoder output $\hat{M}_{dec}(t)$ (or CAM $\hat{M}_{cam}(t)$) and RS $M_{rs}(t)$ through \eqref{recurseed_seg} (or \eqref{enc_rec_loss}), leading to indirect (or direct) improvements in the CAM $\hat{M}_{cam}(t)$ in the next step as the latent feature of the decoder. For stable training, CF is applied to generate the pseudo label $M^t_{cl}=CF(M_{rs}(t))$ in \eqref{certainfilter} and to avoid the influence of uncertain labels. Through this process, CF filters certain and uncertain regions of the RS, thereby enhancing the reliability of the learning.

\subsection{EdgePredictMix}
\label{section:edgepredictmix}

\begin{figure}[t]%
    \centering%
    \includegraphics[width=1.0\linewidth, height=0.85\linewidth]{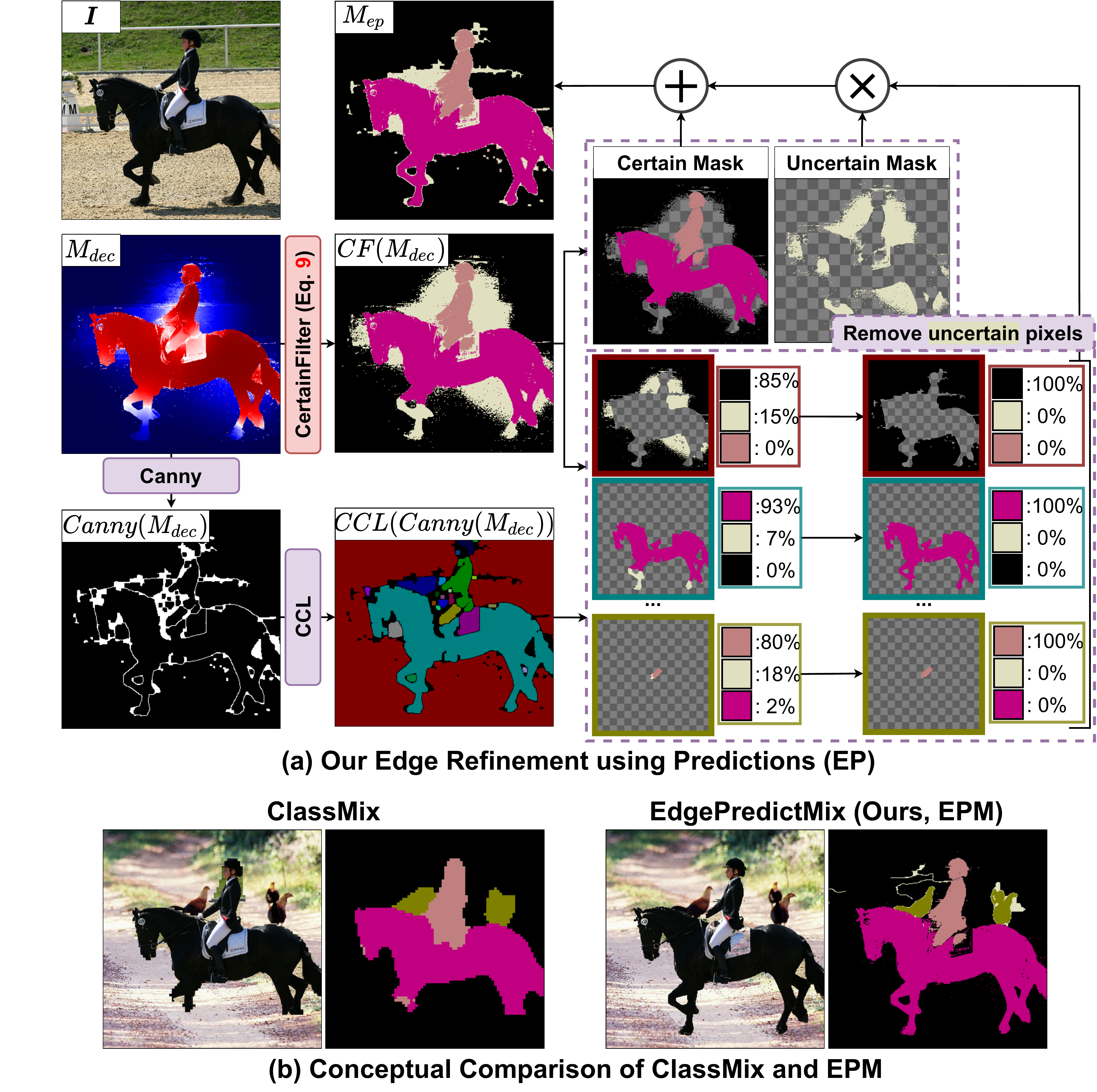}
    \vspace{-0.7cm}
    \caption{
        (a) Illustration of EP. {We create a superpixel separated from the edge between class probabilities and then vote the major class within each superpixel label to remove uncertain pixels. (b) Comparison of recent mixing method (ClassMix). Our EP eliminates uncertain regions in pseudo masks by superpixel-based relabeling, significantly improving WSSS performance.}
    }
    \label{fig:ep}
    \vspace{-0.7cm}
\end{figure}

{We introduce EPM, a novel data augmentation method designed to address IBDA-WSSS-IL. EPM entails two primary components: mask refinement and mask synthesis. In the refinement phase, termed EP, we utilize \textit{absolute and relative per-pixel probability values}. Specifically, we threshold the absolute class probabilities of each pixel to identify uncertain regions. Concurrently, we extract an edge from the mask by assessing the relative differences between adjacent per-pixel probability values and produce superpixels from the edge. Uncertain pixels with low absolute probabilities are then complemented by selecting the predominant class within each superpixel. For edge and superpixel extraction, we employ established algorithms, Canny \cite{canny1986computational} and Connected-component labeling (CCL) \cite{rosenfeld1966sequential}. Two EP-refined masks with their associated original images are merged in the synthesis phase. The EPM process is depicted in Fig. \ref{fig:ep} and formulated in the following steps for two arbitrary indices $i,j \in \{1:B\}$ in a mini-batch.}

The first step of EPM is to supplement the uncertain mask produced from the model by EP so that the decoder output ${M}^i_{dec}(t)$ in the current epoch $t$ is converted to ${M}^i_{ep}(t)$ as in \eqref{ep_alg}. Then, we extract the union of all EP-refined foregrounds for image $I_i$ as $\mathcal{M}^i_{fg}$ in \eqref{mask_fg}.  
\begin{small}
\begin{align}\nonumber 
& {M}^i_{dec}(t) = PAMR\Big(D_{\theta^{dec}_{t}}(E_{\theta^{enc}_{t}}(I_{i}));\mathcal{W}\Big) \\\label{ep_alg}
& {M}^i_{ep}(t) =  EP({M}^i_{dec}(t)) \\\nonumber 
& \mathcal{R}^i_{c}(t) = \{ (k,n) \, | \, {M}^i_{ep}(t)[c,k,n] > \delta_{fg} \} \\\label{mask_fg}
&\mathcal{M}^i_{fg} =  \mathbbm{1} \big[ {\cup}_{c \in \mathcal{C}} \,  \mathcal{R}^i_{c}(t) \big],
\end{align}
\end{small}
{where $\mathbbm{1}$ denotes the element-wise indicator operator and {$\mathcal{W}$ is set as the active region of $D_{\theta^{dec}_{t}}(E_{\theta^{enc}_{t}}(I_{i}))$}.} The second step is pasting certain regions  $\mathcal{M}^i_{fg}$ of an image $I_i$ and its corresponding EP-refined-mask set ${M}^i_{ep}(t)$ (or its RS seed as encoder output ${M}^i_{rs}(t)$) onto the image $I_j$ and its EP-refined-mask set ${M}^j_{ep}(t)$ (or its RS seed ${M}^j_{rs}(t)$), respectively, as shown in $I_{i\rightarrow j}$ in \eqref{image_ij} and $M^{rs}_{i\rightarrow j}(t)$ in \eqref{mask_ij} (or $M^{cam}_{i\rightarrow j}(t)$ in \eqref{dec_mask_ij}). In other words, $I_{i\rightarrow j}$, $M^{seg}_{i\rightarrow j}(t)$, and $M^{rs}_{i\rightarrow j}(t)$ correspond to the (EP-based) synthesizing results in the original image, decoder, and encoder, respectively. 
\begin{small}
\begin{align}\label{image_ij} 
&I_{i\rightarrow j} =  I_{i} \odot \mathcal{M}^i_{fg} + I_{j} \odot (1-\mathcal{M}^i_{fg})\\\label{mask_ij} 
&M^{seg}_{i\rightarrow j}(t) = {M}^i_{ep}(t) \odot \mathcal{M}^i_{fg} + {M}^j_{ep}(t) \odot (1-\mathcal{M}^i_{fg})\\\label{dec_mask_ij} 
&M^{rs}_{i\rightarrow j}(t) = {M}^i_{rs}(t) \odot \mathcal{M}^i_{fg} + {M}^j_{rs}(t) \odot (1-\mathcal{M}^i_{fg}) 
\end{align}
\end{small} 
We then make the network train mixed images and labels as follows:
\begin{small}
\begin{align}\nonumber 
\theta_{t}   =& \\\nonumber 
\theta_{t-1}  
&- (*) 
\\\label{mix_seg} 
&- \eta \frac{\partial}{\partial{\theta}}\mathbb{E}_{\boldsymbol{I}}\big[ \mathcal{L}^{mix}_{seg} \big(\hat{M}^{mix}_{dec}(\tau),M^{mix}_{seg}({\tau}); \theta \big)\big]\Big|_{\theta = \theta_{{\tau}},\tau=t-1},
\\\label{mix_rec} 
&- \eta \frac{\partial}{\partial{\theta}}\mathbb{E}_{\boldsymbol{I}}\big[ \mathcal{L}^{mix}_{rec} \big(\hat{M}^{mix}_{cam}(\tau),M^{mix}_{rs}({\tau}); \theta \big)\big]\Big|_{\theta = \theta_{{\tau}},\tau=t-1},
\vspace{-0.5cm}
\end{align}
\end{small}
where $(*)$ refers to all terms in  \eqref{recurseed_label}, \eqref{recurseed_seg}, and \eqref{enc_rec_loss}, 
\begin{small} 
\begin{align}\nonumber 
\boldsymbol{I}_{mix} &=  \big\{I_{i \rightarrow j}\, \big| \, j \sim  \textup{unif}(\{1:B\}/i) \textup{ for $i \in \{1:B\}$}\big\}  \\\label{m_mix}
M^{mix}_{seg}(t) &=  \big\{M^{seg}_{i \rightarrow j}(t)\, \big| I_{i \rightarrow j} \in \boldsymbol{I}_{mix} \big\}  
\\\label{m_mix_pred}
\hat{M}^{mix}_{dec}(t) &= D_{\theta^{dec}_{t}}(E_{\theta^{cam}_{t}}(\boldsymbol{I}_{mix}))
\\\label{m_mix_dec}
M^{mix}_{rs}(t) &=  \big\{M^{rs}_{i \rightarrow j}(t)\, \big| I_{i \rightarrow j} \in \boldsymbol{I}_{mix} \big\}   
\\\label{y_mix} 
\hat{M}^{mix}_{cam}(t) &= A_{\theta^{cls}_{t}}(E_{\theta^{cam}_{t}}(\boldsymbol{I}_{mix}))
\end{align}
\end{small}
Using the proposed EPM, we obtain other mini-batch $\boldsymbol{I}_{mix}$ of $B$ mixed images by selecting each sample in the mini-batch $\boldsymbol{I}$, choosing another sample randomly from the remaining samples, and mixing them. The pseudo masks and the decoder outputs for $\boldsymbol{I}_{mix}$ at epoch $t$ are given by $M^{mix}_{seg}(t)$ in \eqref{m_mix} and $\hat{M}^{mix}_{dec}(t)$ in \eqref{m_mix_pred}, respectively, through additional optimizations specified as follows: The first loss term in \eqref{mix_seg} (or second loss term in \eqref{mix_rec}) corresponds to the cross-entropy loss (or the reconstruction loss) that makes the decoder output $\hat{M}^{mix}_{dec}(t)$ (or the encoder output $\hat{M}^{mix}_{cam}(t)$) reproduce the mixed mask refined by EP $M^{mix}_{seg}(t)$ (or the mixed RS $M^{mix}_{rs}(t)$).

\section{Experiments}
\label{section:exp}

\subsection{Reproducibility}

\paragraph{Details for datasets.} 

The PASCAL VOC 2012 dataset \cite{everingham2010pascal} comprises 21 classes, which include a background class, and is divided into three subsets: training (1,464 images), validation (1,449 images), and test (1,456 images). Following the standard WSSS setup, \emph{e.g}., \cite{ahn2018learning,wang2020self}, we use an augmented training dataset with 10,582 images. The MS COCO 2014 dataset \cite{lin2014microsoft} consists of 81 classes with 82,783 training and 40,504 validation images. The outcomes for the PASCAL VOC 2012 validation and test sets are sourced from the official evaluation server.

\paragraph{Implementation details.} 

Our single-stage learning framework (SLF) is integrated with a random walk (RW) \cite{ahn2019weakly} for a multi-stage learning framework. We incorporate our CertainFilter in \eqref{certainfilter} with $\delta_{fg} = 0.55$ and $\delta_{bg} = 0.10$. Following \cite{araslanov2020single}, we employ ResNet-50 (R50) with DeepLabv3+ \cite{chen2018encoder} for our SLF and follow a multi-scale strategy with CRF \cite{krahenbuhl2011efficient} during testing. We use a stochastic gradient descent optimizer with a weight decay of $4e^{-5}$. The initial learning rate is set to 0.1, decreasing polynomially at 0.9. Image augmentations involve random scaling, horizontal flips, and cropping to $512 \times 512$ resolution. We conduct all experiments on a single RTX A6000 GPU using PyTorch, and our training time for PASCAL VOC 2012 concludes in under 24 hours.

\begin{table}[t]
    \centering
    \caption{ 
    Performance comparison of WSSS methods in terms of mIoU ($\%$) on PASCAL VOC 2012 and COCO 2014. * indicates the backbone of VGG-16. Sup., supervision; $\mathcal{I}$, image-level class labels; $\mathcal{S}$, the saliency supervision; $\mathcal{D}$, an additional dataset; $\mathcal{B}$, box labels; $\mathcal{P}$, point labels. 
  }
  \vspace{-0.10cm}
  \begin{scriptsize}
  \begin{tabular}{p{0.195\textwidth} c p{0.02\textwidth} p{0.015\textwidth} p{0.015\textwidth} c}
    \toprule
    \multirow{2}{*}{\begin{tabular}[c]{@{}c@{}}Method\end{tabular}} & \multirow{2}{*}{\begin{tabular}[c]{@{}c@{}}Backbone\end{tabular}} & \multirow{2}{*}{\begin{tabular}[c]{@{}c@{}}Sup.\end{tabular}} & \multicolumn{2}{c}{VOC}  & COCO \\
           &          &      & \emph{val} & \emph{test} & \emph{val} \\
    \hline 
    \multicolumn{6}{c}{\textbf{Single-stage learning framework using image-level supervision only}} \\[1.5pt]
    \hline
    RRM \cite{zhang2020reliability} & WR38 & $\mathcal{I}$ & 62.6 & 62.9 & - \\
    SSSS \cite{araslanov2020single} & WR38 & $\mathcal{I}$ & 62.7 & 64.3 & - \\
    AFA \cite{ru2022learning} & MiT-B1 & $\mathcal{I}$ & 66.0 & 66.3 & 38.9 \\
    ToCo \cite{ru2023token} & ViT-B & $\mathcal{I}$ & \textbf{69.8} & 70.5 & 41.3 \\
    \rowcolor{maroon!25}
    RS (Ours, single-stage) & R50 & $\mathcal{I}$ & 66.5 & 67.9 & 40.0 \\ 
    \rowcolor{maroon!25}
    RSEPM (Ours, single-stage) & R50 & $\mathcal{I}$ & 69.5 & \textbf{70.6} & \textbf{42.2} \\
    \hline 
    \multicolumn{6}{c}{\textbf{Multi-stage learning framework using image-level supervision only}} \\[1.5pt]
    \hline
    PSA \cite{ahn2018learning} & WR38 & $\mathcal{I}$ & 61.7 & 63.7 & - \\
    IRNet \cite{ahn2019weakly} & R50 & $\mathcal{I}$ & 63.5 & 64.8 & - \\
    SEAM \cite{wang2020self} & WR38 & $\mathcal{I}$ & 64.5 & 65.7 & 31.9 \\
    CONTA \cite{zhang2020causal} & R101 & $\mathcal{I}$ & 66.1 & 66.7 & 32.8 \\
    CDA \cite{su2021context} & WR38 & $\mathcal{I}$ & 66.1 & 66.8 & 33.2 \\
    AdvCAM \cite{lee2021anti} & R101 & $\mathcal{I}$ & 68.1 & 68.0 & - \\
    RIB \cite{lee2021reducing} & R101 & $\mathcal{I}$ & 68.3 & 68.6 & 43.8 \\
    ReCAM \cite{chen2022class} & R101 & $\mathcal{I}$ & 68.5 & 68.4 & - \\
    ADEHE \cite{liu2022adaptive} & R101 & $\mathcal{I}$ & 68.6 & 68.9 & - \\
    AMR \cite{qin2022activation} & R101 & $\mathcal{I}$ & 68.8 & 69.1 & - \\
    SIPE \cite{chen2022self} & R101 & $\mathcal{I}$ & 68.8 & 69.7 & 40.6 \\
    AMN \cite{lee2022threshold} & R101 & $\mathcal{I}$ & 69.5 & 69.6 & 44.7 \\
    MCTformer \cite{xu2022multi} & WR38 & $\mathcal{I}$ & 71.9 & 71.6 & 42.0 \\
    SANCE \cite{li2022towards} & R101 & $\mathcal{I}$ & 70.9 & 72.2 & 44.7$\dagger$ \\
    BECO \cite{rong2023boundary} & R101 & $\mathcal{I}$ & 72.1 & 71.8 & - \\
    ACR \cite{kweon2023weakly} & WR38 & $\mathcal{I}$ & 71.9 & 71.9 & 45.3 \\
    OCR \cite{cheng2023out} & WR38 & $\mathcal{I}$ & 72.7 & 72.0 & 42.5 \\
    \rowcolor{maroon!25}
    RS (Ours, multi-stage) & R101 & $\mathcal{I}$ & 72.8 & 72.8 & 45.8 \\ 
    \rowcolor{maroon!25}
    RSEPM (Ours, multi-stage) & R101 & $\mathcal{I}$ & \textbf{74.4} & \textbf{73.6} & \textbf{46.4} \\
    \hline
    \multicolumn{6}{c}{\textbf{Multi-stage learning framework using additional supervision}} \\[1.5pt]
    \hline
    DSRG \cite{huang2018weakly} & R101 & $\mathcal{I}$+$\mathcal{S}$ & 61.4 & 63.2 & 26.0* \\
    FickleNet \cite{lee2019ficklenet} & R101 & $\mathcal{I}$+$\mathcal{S}$ & 64.9 & 65.3 & - \\
    EDAM \cite{wu2021embedded} & R101 & $\mathcal{I}$+$\mathcal{S}$ & 70.9 & 70.6 & - \\
    CLIMS \cite{xie2022clims} & R50 & $\mathcal{I}$+$\mathcal{D}$ & 69.3 & 68.7 & - \\
    W-OoD \cite{lee2022weakly} & R101 & $\mathcal{I}$+$\mathcal{D}$ & 70.7 & 70.1 & - \\
    L2G \cite{jiang2022l2g} & R101 & $\mathcal{I}$+$\mathcal{S}$ & 72.1 & 71.7 & 44.2 \\
    RCA \cite{zhou2022regional} & R101 & $\mathcal{I}$+$\mathcal{S}$ & 72.2 & 72.8 & 36.8* \\
    PPC \cite{du2022weakly} & R101 & $\mathcal{I}$+$\mathcal{S}$ & 72.6 & 73.6 & - \\
    \hline
    \multicolumn{6}{c}{\textbf{Multi-stage learning framework using different supervision}} \\[1.5pt]
    \hline
    TEL \cite{liang2022tree} & R101 & $\mathcal{P}$ & 74.2 & - & - \\
    BAP \cite{oh2021background} & R101 & $\mathcal{B}$ & 79.2 & - & - \\
    \bottomrule
  \end{tabular} 
  \label{tab:comp_voc_coco}
  \end{scriptsize}
  \vspace{-0.30cm}
\end{table}

\subsection{Comparison with state-of-the-art approaches}
\label{ssec:results}


Under the same supervision, our SLF method outperforms most methods, even if only RecurSeed (RS) is used, on PASCAL VOC 2012 and MS COCO 2014 datasets. The Integrating our EdgePredictMix (EPM) further elevates the performance compared to employing our RecurSeed (RS) only, proving the effectiveness of our RS and EPM.



In contrast with PASCAL VOC 2012, MS COCO 2014 is more challenging because it has four times categories and requires to detection of non-salient foreground objects. As shown in the last column of Tab. \ref{tab:comp_voc_coco}, even for MS COCO 2014, our single- and multi-stage methods achieve new state-of-the-art mIoUs of $42.2\%$ and $46.4\%$, marking improvements of $0.9\%$ and $1.1\%$ over recent studies, respectively. Notably, on the PASCAL VOC 2012 \emph{test} set, our single-stage method with fewer parameters (\emph{i.e.}, R50) outperforms most multi-stage methods using the same supervision. This result underscores that the latest performance can be achieved only with a single-stage method.

\subsection{Analysis}
\label{section:disc}


\begin{table}
  \centering
  \caption{ 
    Element-wise component analysis on PASCAL VOC 2012 $train$ set. With RS, SCG and SCG+PAMR denote $M_{rs}(t)$ in \eqref{mrs} by removing and preserving operator $PAMR(\cdot)$ throughout the training, respectively. RS indicates $t=1$ ( ) or $30$ (\checkmark). $*$ denotes the decoder map result.
  }
  \begin{scriptsize}
  \begin{tabular}{p{0.060\textwidth} p{0.02\textwidth} p{0.04\textwidth} | p{0.03\textwidth} p{0.095\textwidth} p{0.095\textwidth}}
    \toprule
    \textbf{RS (Ours)} & SCG     & PAMR    & mIoU & FP & FN \\
    \hline 
         &  \checkmark &         & 58.0 & 0.268 & 0.165 \\
         &  \checkmark &  \checkmark & 59.3 & 0.225 (\textcolor{blue}{$\downarrow$ 0.043}) & 0.194 (\textcolor{red}{$\uparrow$ 0.029})\\
    \hline 
     \checkmark &  \checkmark &       & 65.2 & 0.216 & 0.143 \\
     \rowcolor{maroon!25}
     \checkmark &  \checkmark &  \checkmark & 65.9 & 0.210 (\textcolor{blue}{$\downarrow$ 0.006}) & 0.141 (\textcolor{blue}{$\downarrow$ 0.002})\\
    \hline 
     \checkmark &  \checkmark &      & *67.4 & *0.196 & *0.141 \\
     \rowcolor{maroon!25}
     \checkmark &  \checkmark &  \checkmark & *70.7 & *0.171 (\textcolor{blue}{$\downarrow$ 0.025}) & *0.134 (\textcolor{blue}{$\downarrow$ 0.007}) \\
    \bottomrule
  \end{tabular}
  \label{tab:rs}
    \end{scriptsize}
    \vspace{-0.1cm}  
\end{table}

\paragraph{Novelty of RS.} Existing SCG and PAMR studies, which use a fully trained network, were individually developed for the post-processing method on the CAM. As shown in Tab. \ref{tab:rs}, the simple integration of SCG and PAMR (\emph{i.e.}, w. SCG and PAMR but w.o. RS) not only shows a marginal improvement but also fails to reduce FP and FN simultaneously. By contrast, our RS decreases both FP and FN by leveraging the recursive update, achieving substantial improvements (+11.4\%) to the simple integration. These results indicate that RS makes both SCG and PAMR utilized as recursively together to mitigate FPN-WSSS-IL. Also, as shown in Tab. \ref{tab:ordering}, we demonstrate the effect of the second-order correlation in recursion. With RS, the results from the encoder and decoder yield significant improvements (+7.4\%) versus the first-order correlation. However, the second-order correlation without RS improves the first-order correlation marginally. Therefore, we verify the validity of the second-order correlation in the recursion principle, \emph{i.e.}, the novelty of the proposed RS.

\begin{table}
  \centering
  \caption{ 
    Effect of the second-order correlation on PASCAL VOC 2012 $train$ set. All experiments are based on applying PAMR. RS indicates $t=1$ ( ) or $30$ (\checkmark). $*$ denotes the decoder map result.
  }
  \vspace{-0.10cm}
  \begin{scriptsize}
  \begin{tabular}{p{0.04\textwidth} p{0.125\textwidth} p{0.135\textwidth} | p{0.09\textwidth} }
    \toprule
    \textbf{RS (Ours)} & first-order correlation & second-order correlation & mIoU \\ 
    \hline 
                &  \checkmark &             & 57.6 \\
                &  \checkmark &  \checkmark & 59.3 (+1.7\%) \\
    \hline
     \checkmark &  \checkmark &             & 60.8 \\
     \rowcolor{maroon!25}
     \checkmark &  \checkmark &  \checkmark & \textbf{65.9 (+5.1\%)} \\
    \hline 
     \checkmark &  \checkmark &      & *63.3 \\
     \rowcolor{maroon!25}
     \checkmark &  \checkmark &  \checkmark & \textbf{*70.7 (+7.4\%)} \\
    \bottomrule
  \end{tabular}
  \label{tab:ordering}
  \end{scriptsize}
  \vspace{-0.3cm}  
\end{table}

\paragraph{Novelty of EPM.} The previous data augmentation (DA) methods rely on unsuitable seeds (\emph{e.g.}, CAM) or ideal annotations (\emph{e.g.}, pixel-wise annotations). Nevertheless, we apply our RS in our EPM and other DA methods for a fair comparison. In Tab. \ref{tab:mix}, CutMix \cite{yun2019cutmix}, Saliency Grafting \cite{park2022saliency}, and CDA \cite{su2021context} decrease approximately 2\% of mIoU. ClassMix \cite{olsson2021classmix} only shows a marginal improvement for the WSSS due to mixing predicted masks without mask refinement. Our EPM not only trains mixed masks (like ClassMix) but also further refines pseudo masks by leveraging the reconstituted edge from uncertain regions, thereby resulting in the highest WSSS performance (mIoU $75.2\%$) compared to other DA techniques (the best mIoU $71.2\%$).

\begin{table}
  \vspace{-0.20cm}
  \caption{
    Performance comparison with the proposed EPM and existing mixing methods in terms of mIoU (\%) on PASCAL VOC 2012 $train$ set. * denotes our implementation for a fair comparison.
  } 
  \vspace{-0.10cm}
  \centering
  \begin{scriptsize} 
  \begin{tabular}{p{0.29\textwidth} | p{0.05\textwidth} p{0.08\textwidth} }
    \toprule
    Method & Backbone & mIoU \\
    \hline 
    RecurSeed & R50 & 70.7 \\
    RecurSeed + *CutMix \cite{yun2019cutmix} & R50 & 68.5 \\
    RecurSeed + *Saliency Grafting \cite{park2022saliency} & R50 & 68.6 \\
    RecurSeed + *CDA \cite{su2021context} & R50 & 69.0 \\
    RecurSeed + *ClassMix \cite{olsson2021classmix} & R50 & 71.2 \\
    \rowcolor{maroon!25}
    RecurSeed + EdgePredictMix (Ours) & R50 & \textbf{75.2} \\
    \bottomrule
  \end{tabular}
  \label{tab:mix}
  \end{scriptsize}
  \vspace{-0.15cm}
\end{table}

\begin{table}
  \centering
  \caption{ 
    Effect of our EPM with other single-stage methods on PASCAL VOC 2012 $val$ set. 
  }
  \vspace{-0.10cm}
  \begin{scriptsize}
  \begin{tabular}{p{0.29\textwidth} | p{0.05\textwidth} p{0.08\textwidth} }
    \toprule
    Method & Backbone & mIoU \\ 
    \hline 
    RRM \cite{zhang2020reliability} & WR38 & 62.6 \\
    \rowcolor{maroon!25}
    + EdgePredictMix (Ours) & WR38 & \textbf{65.3 (+2.7\%)} \\
    \hline
    SSSS \cite{araslanov2020single} & WR38 & 62.7 \\
    \rowcolor{maroon!25}
    + EdgePredictMix (Ours) & WR38 & \textbf{65.5 (+2.8\%)} \\
    \hline
    AFA \cite{ru2022learning} & MiT-B1 & 66.0 \\
    \rowcolor{maroon!25}
    + EdgePredictMix (Ours) & MiT-B1 & \textbf{68.5 (+2.5\%)} \\
    \bottomrule
  \end{tabular}
  \label{tab:epm}
    \end{scriptsize}
    \vspace{-0.30cm}
\end{table}

Similar to CutMix \cite{yun2019cutmix}, our EPM can be applied to any WSSS methods, as in Tab. \ref{tab:epm}. We use Canny \cite{canny1986computational} as it is the well-known method of edge detection. Therefore, we replace it with other methods \cite{xie2015holistically}; our method shows consistent superiority regardless of edge detection methods on the PASCAL VOC 2012 \emph{val} set. (RSEPM using Canny \cite{canny1986computational}: 69.5\%, RSEPM using HED \cite{xie2015holistically}: 69.4\%). 

\vspace{-0.3cm}
 
\paragraph{Importance of the proposed components.} To clarify the effect of the proposed RS and EPM, we conduct ablation studies related to loss functions and EP on our single-stage network. In Tab. \ref{tab:loss}, we report different combinations as we add or remove components. First, as shown in row 3, we observe that training both the encoder and decoder with our RS significantly improves the decoder's performance and brings an explicit gain for the CAM as the recursively advanced features lead to better pseudo masks in the next step. In addition, RS with EP achieves 71.7\% of the decoder in row 4. The result shows that EP itself achieves meaningful improvement even without mixing augmentation. Second, in rows 5 and 7, training mixed results without EP shows a certain level of improvement and outperforms ClassMix \cite{olsson2021classmix} as the model only trains reliable labels from CF. We present samples of qualitative segmentation results on PASCAL VOC 2012 and MS COCO 2014 sets in Appendix C.2. In the last row, we achieve the best performance when applying essential components together with EP by exploiting absolute and relative per-pixel probability information. Therefore, we demonstrate the novelty and validity of the proposed components (\emph{i.e.}, RS and EPM).

\begin{table}
    \centering
  \caption{ 
    Effect of key components for the proposed RS and EPM in terms of mIoU (\%) on PASCAL VOC 2012 $train$ set.  
  }
  \vspace{-0.20cm}
  \begin{scriptsize}
  \begin{tabular}{p{0.001\textwidth} c c c | c c c | c c}
    \toprule
    & $L_{cls}$  & $L_{seg}$  & $L_{rec}$  & $L^{mix}_{seg}$ & $L^{mix}_{rec}$ & EP         & CAM  & Decoder \\
    \hline 
      & \multicolumn{3}{c|}{\textbf{RecurSeed}}  & \multicolumn{3}{c|}{\textbf{EdgePredictMix}} & & \\
    1 &  \checkmark &       &       &           &            &       & 46.9 & 17.4    \\
    2 &  \checkmark &  \checkmark &             &            &            &       & 51.6 & 65.8    \\
    3 &  \checkmark &  \checkmark &  \checkmark &            &            &       & 57.9 & 70.7    \\
    4 &  \checkmark &  \checkmark &  \checkmark &            &            & \checkmark & 59.0 & 71.7    \\
    \hline
    5 &  \checkmark &  \checkmark &  \checkmark &  \checkmark      &            &       & 61.0 & 73.5    \\
    6 &  \checkmark &  \checkmark &  \checkmark &  \checkmark      &            &  \checkmark & 60.8 & 74.5    \\
    7 &  \checkmark &  \checkmark &  \checkmark &  \checkmark      &  \checkmark      &       & 61.9 & 73.6    \\
    \rowcolor{maroon!25}
    8 &  \checkmark &  \checkmark &  \checkmark &  \checkmark      &  \checkmark      &  \checkmark & \textbf{63.4} & \textbf{75.2}    \\
    \bottomrule
  \end{tabular}
  \label{tab:loss}
    \end{scriptsize}
    \vspace{-0.15cm}
\end{table}

\paragraph{Additional comparison with existing MLFs using RW.}   
In Tab. \ref{tab:rw}, we compare the performance of the proposed SLF with existing MLFs \cite{wang2020self,ahn2019weakly,lee2021reducing} by extending it to MLF with the same configuration of RW. Surprisingly, in all cases before or after the application of RW, our method outperforms the related works by achieving mIoUs of 75.2\% and 76.7\% without and with RW, respectively, exhibiting the performance improvement of at least 15\% and 6\% compared to the prior arts (CPN 57.4\% and RIB 70.6\% respectively). This also proves the superiority of the proposed RS and EPM.  

\begin{table}
  \caption{
    mIoUs ($\%$) of single-stage results produced from a trained network (Seed) and pseudo masks generated by RW \cite{ahn2019weakly} on PASCAL VOC 2012 $train$ set. 
  }
  \vspace{-0.20cm}
  \centering
  \begin{scriptsize} 
  \begin{tabular}{p{0.235\textwidth} c c c}
    \toprule
    Method & Backbone & Seed & Seed + RW \\
    \hline 
    SEAM \cite{wang2020self} & WR38 & 55.4 & 63.6 \\
    IRNet \cite{ahn2019weakly} & R50 & 48.8 & 66.3 \\
    CDA \cite{su2021context} & R50 & 50.8 & 67.7 \\
    CPN \cite{zhang2021complementary} & WR38 & 57.4 & 67.8 \\
    CONTA \cite{zhang2020causal} & R50 & 48.8 & 67.9 \\
    AdvCAM \cite{lee2021anti} & R50 & 55.6 & 69.9 \\
    PPC \cite{du2022weakly} & WR38 & 61.5 & 70.1 \\
    RIB \cite{lee2021reducing} & R50 & 56.5 & 70.6 \\ 
    \rowcolor{maroon!25}
    RS (Ours) & R50 & 70.7 & 74.8 \\
    \rowcolor{maroon!25}
    RSEPM (Ours) & R50 & \textbf{75.2} & \textbf{76.7} \\
    \bottomrule
  \end{tabular}
  \label{tab:rw}
  \end{scriptsize}
  \vspace{-0.30cm}
\end{table}

\begin{figure*}[t]%
    \centering 
    \includegraphics[width=1.0\linewidth, height=0.15\linewidth]{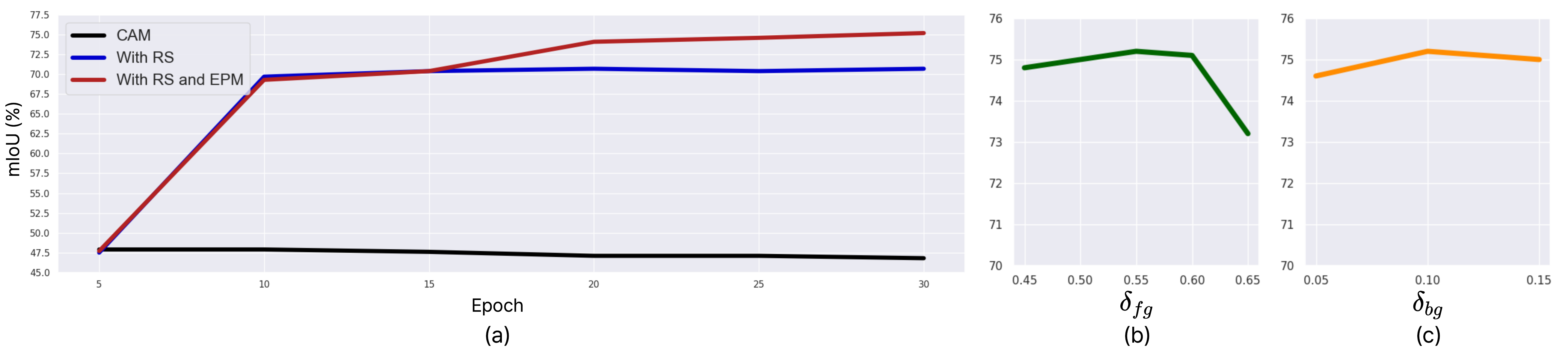}
    \vspace{-0.45cm}
    \caption{
        (a) Effects of RS and EPM per epoch.
        (b) Effect of foreground threshold $\delta_{fg}$.
        (c) Effect of background threshold $\delta_{bg}$.
    }
    \label{fig:training}
    \vspace{-0.50cm}
\end{figure*}

\paragraph{Hyperparameter analysis.}
To validate the effects of the proposed RS and EPM, we evaluate our single-stage results for each epoch, as shown in Fig. \ref{fig:training}(a). Initially, the mIoU increases steeply through the RS between the 5th and 10th epochs. However, the mIoU then saturates between the 10th and 15th epochs. To address this saturation, we apply EPM after the 15th epoch. Here, $\delta_{fg}$ and $\delta_{bg}$ in \eqref{certainfilter} control the foreground and background regions. Although we observe a sufficiently high mIoU above a certain level irrespective of changes of $\delta_{fg}$ and $\delta_{bg}$ in Figs \ref{fig:training}(b)-(c), the best performance can be found through their adjustment, proving the validity of two thresholds.

\vspace{-0.30cm}

\begin{figure}[t]%
    \centering%
    \includegraphics[width=1.0\linewidth, height=0.8\linewidth]{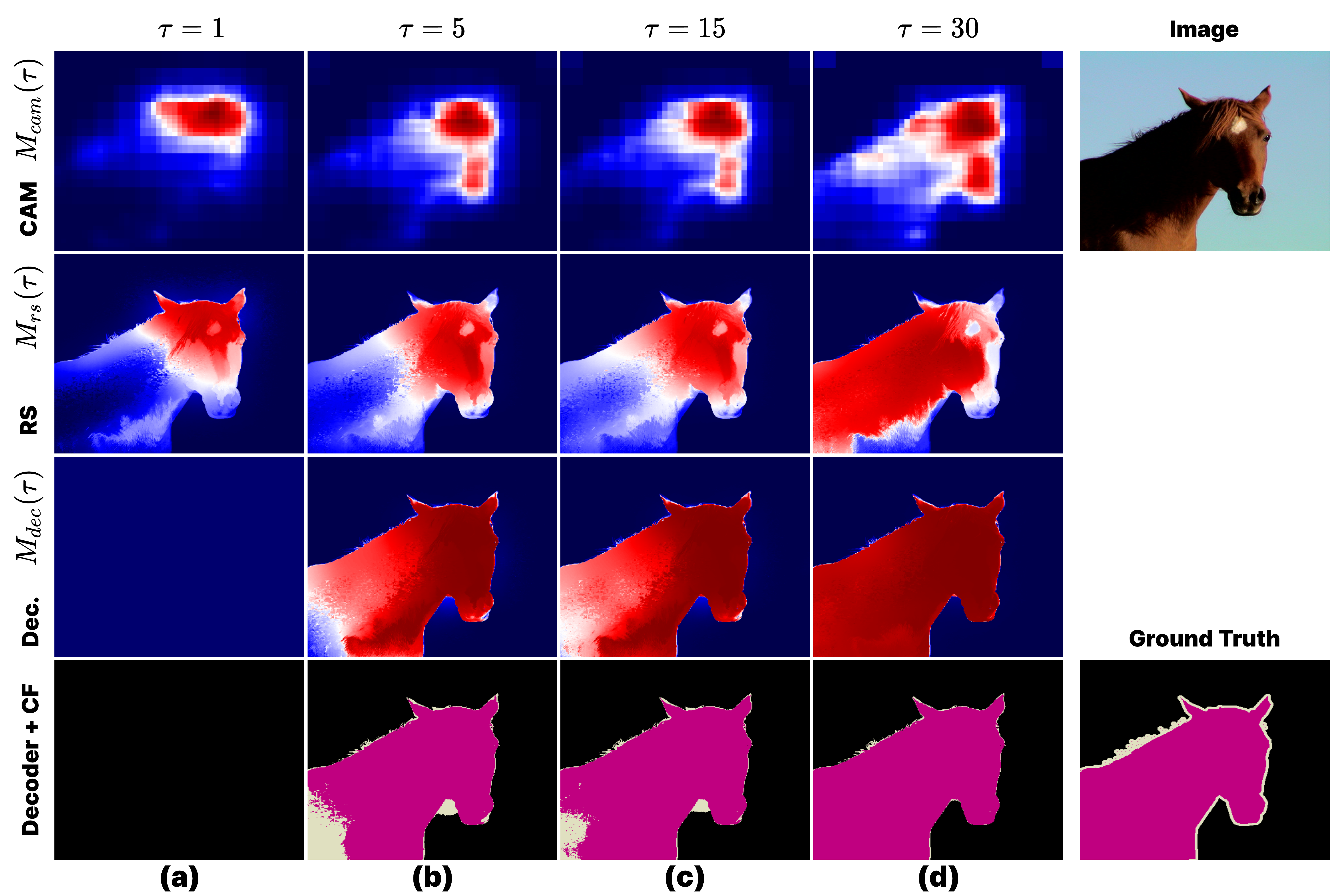}
    \vspace{-0.70cm}
    \caption{
        Visualization of attention maps and pseudo masks with recursive improvements.
    }
    \label{fig:impro}
    \vspace{-0.40cm}
\end{figure}

\paragraph{Qualitative improvements.}
Fig. \ref{fig:impro} illustrates the results produced by our single-stage method (including CAM, RS, and Decoder). We then generate pseudo masks from the decoder output by using CF. For the initial training step shown in Fig. \ref{fig:impro}(a), CAM and SCG cover the most discriminative part of an object. In addition, the decoder output from the segmentation branch fails to detect all of the foregrounds owing to insufficient training steps. After we apply RS to train the segmentation branch (Figs \ref{fig:impro}(b)-(c)), the predicted masks are better than those at the initial step ($t = 1$). As the learning steps progress through EPM (Fig. \ref{fig:impro}(d)), the CAM and SCG represent more integral regions of an object, and the segmentation branch produces an accurate seed. Specifically, we secure the data diversity and increase the mIoU by more than 5\%. The qualities of pseudo masks (Decoder+CF) are improved progressively in Figs \ref{fig:impro}(a)-(d), and the final mask in Fig. \ref{fig:impro}(d) is then close to the ground truth.


\vspace{-0.40cm}

\begin{figure}[t]%
    \centering%
    \includegraphics[width=1.0\linewidth, height=0.80\linewidth]{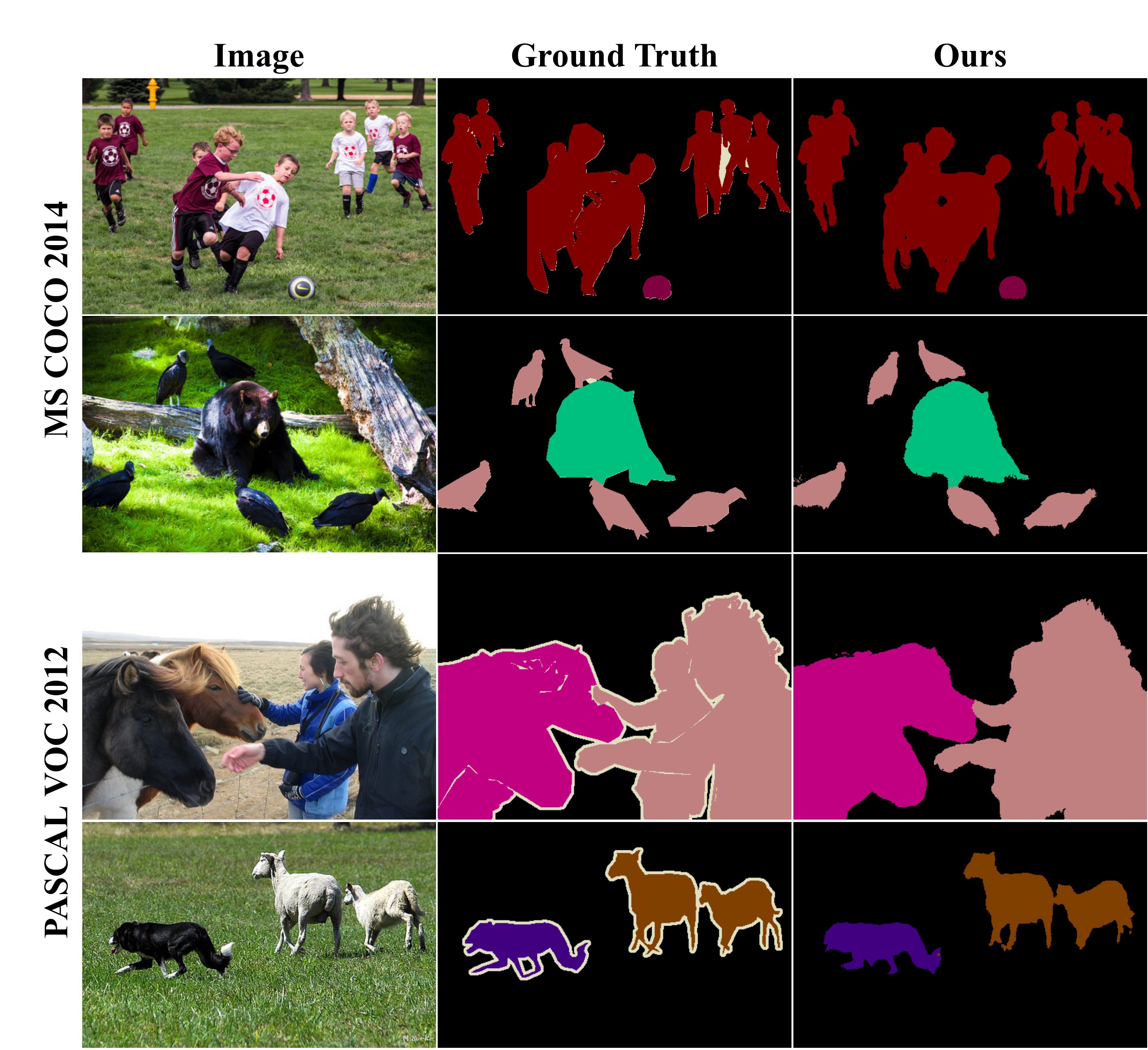}
    \vspace{-0.60cm}
    \caption{
        Qualitative results on the VOC and COCO \emph{val} sets.
    }
    \label{fig:qualitative_results}
    \vspace{-0.25cm}
\end{figure}

\paragraph{Qualitative results.} 
We present some examples of qualitative segmentation results produced by our method on both the PASCAL VOC 2012 and MS COCO 2014 $val$ sets in Fig. \ref{fig:qualitative_results}, respectively. These results show that our method not only performs well for different complex scenes, small objects, or multiple instances but also can achieve a satisfactory segmentation performance for various challenging scenes. 

\paragraph{Complexity.} Tab. \ref{tab:complexity} shows our efficiency. Ours outperforms previous studies while using fewer parameters. The proposed RS and EPM are only used for training, so our method does not require extra complexity at inference time. 

\vspace{-0.20cm}

\begin{table}
  \centering
  \caption{ 
    Comparison with ours and other single-stage methods on PASCAL VOC 2012 $test$ set.
  }
  \vspace{-0.20cm}
  \begin{scriptsize}
  \begin{tabular}{p{0.15\textwidth} | p{0.05\textwidth} p{0.07\textwidth} p{0.06\textwidth} p{0.03\textwidth} }
    \toprule
    Method & Backbone & Params (M) & MACs (G) & mIoU \\ 
    \hline 
    RRM \cite{zhang2020reliability} & WR38 & 124 & 3990 & 62.9 \\
    SSSS \cite{araslanov2020single} & WR38 & 137 & 678 & 64.3 \\
    AFA \cite{ru2022learning} & MiT-B1 & 14 & 52 & 66.3 \\
    ToCo \cite{ru2023token} & ViT-B & 106 & 250 & 70.5 \\
    \rowcolor{maroon!25}
    RSEPM (Ours) & R50 & 40 & 32 & \textbf{70.6} \\
    \bottomrule
  \end{tabular}
  \label{tab:complexity}
  \end{scriptsize}
  \vspace{-0.50cm}
\end{table}


\section{Conclusion}
In this study, we proposed RecurSeed (RS) and EdgePredictMix (EPM) to improve the WSSS-IL performance, achieving new state-of-the-art performances on the PASCAL VOC 2012 and MS COCO 2014 benchmarks. Furthermore, because RS and EPM are learning methods generally applicable to SLFs consisting of an arbitrary encoder and decoder, one can upgrade this backbone to achieve a higher SLF performance. We also expect higher MLF performance can be derived by performing multi-stage extension from the proposed SLF to the latest techniques other than a random walk (RW) or applying our RS and EPM to the individual learning of any recent MLF. From this perspective, we expect that our single- and multi-stage methods have high utility and scalability in weakly or semi-supervised tasks.

\clearpage

\appendix

\section{Background}
\label{section:background}

\subsection{SCG}
\label{section:scg}
The self-correlation map generating module (SCG) \cite{pan2021unveiling} exploits the second-order self-correlation {(SC)} of the feature maps to extract their inherent structural information. SCG uses the first- and second-order SCs $SC^{1}_{l},SC^{2}_{l} \in \mathbb{R}^{hw \times hw}$ for a given feature map $f^l \in \mathbb{R}^{hw \times u}$ of the $i_l$-th layer of the network.
\begin{small}
\begin{align}\nonumber
SC^{1}_{l}[i,j,:,:] &=  \textup{ReLU}\Big(\frac{{f^l_i}^{\top}f^l_j}{\left\|f^l_i\right\|\left\|f^l_j\right\|}\Big), \\\nonumber
SC^{2}_{l}[i,:,:,:] &= {\minmax}_{j} \bigg({\average}_{k}\Big[\frac{{f^l_i}^{\top}f^l_k}{\left\|f^l_i\right\|\left\|f^l_k\right\|} \odot \frac{{f^l_k}^{\top}{f^l_j}}{\left\|f^l_k\right\|\left\|f^l_j\right\|}\Big] \bigg),
\end{align}
\end{small} 
where ${\minmax}_{j}$ denotes the min-max normalization operator along the $j$-th axis, ${\average}_{k}$ denotes the function taking the mean along the $k$-th axis, and $\odot$ denotes the element-wise product.
Then, SCG combines the first- and second-order SCs for L different layers into the following SC, called $HSC$:
\begin{small}
\begin{align}\nonumber 
HSC = \frac{1}{L} \sum_{l = 1}^{L}  \max(SC^{1}_{l},SC^{2}_{l}),
\end{align}
\end{small} 
and applies HSC to obtain the refined CAM as $M_{scg}=SCG(M_{cam})$.
In addition, the proposed RS includes the SCG in the training loop to maximize the effectiveness of the SCG. 
However, the SCG is designed to reduce the FNs of the CAM only at the inference time.
{We additionally reported the effects of the high-order feature correlation for SCG in Appendix.}
\begin{align}\nonumber
M_{scg} &= SCG(M_{cam}) \\\nonumber
&= ReLU(K_{scg}(\{M_{cam}\}_{>\delta_{h}}) - K_{scg}(\{M_{cam}\}_{<\delta_{l}}), 
\end{align}
where $\{M_{cam}\}_{>\delta_{h}}$ denotes the set of ($i,j$) indices satisfying $M_{cam}[i,j]>\delta_{h}$ and $K_{scg}(\{M_{cam}\}_{>\delta_{h}})$ denotes the average activation maps generated from each pixel in $\{M_{cam}\}_{>\delta_{h}}$ based on the HSC,
\vspace{0.1cm}
\begin{small}
\begin{align}\nonumber
K_{scg}(\{M\}_{ \lessgtr \delta }) = \frac{1}{|\{M\}_{\lessgtr\delta }|}\sum_{(i,j) \in \{M\}_{\lessgtr \delta } } HSC[i,j,:,:].
\end{align}
\end{small} 

\subsection{PAMR}
\label{section:pamr}
Pixel-adaptive mask refinement (PAMR) proposed by \cite{araslanov2020single} iteratively refines the CAM $M_{cam}$ as $M_{pamr}= G_{T}(M_{cam})$ by exploiting the image pixel-level affinity matrix $(\alpha_{i,j,k,n})_{(k,n) \in \mathcal{N}(i,j)}$ for $(i,j)\in \mathcal{W}=\{1:h,1:w\}$ of all pixels: 
\begin{small}
\begin{align} 
M_{pamr} &= PAMR(M_{cam};\mathcal{W}) = G_{T}(M_{cam}), 
\end{align}
\end{small} 
where $G_{0}(M_{cam}) = M_{cam}$, $G_{t}(M_{cam})[i,j] = \sum_{(k,n) \in \mathcal{N}(i,j)} \alpha_{i,j,k,n} M_{cam}(k,n)$, $\alpha_{i,j,k,n} = {\exp[\bar{k}(i,j,k,n)]}$ $/ \sum_{(q,r) \in \mathcal{N}(i,j)}\exp[\bar{k}(i,j,q,r)] $. Here, $\bar{k}(\cdot)$ is the average affinity value $k(\cdot)$ across the RGB channels, $k(i,j,k,n)= {-|I_{i,j}-I_{k,n}|}/{\sigma^2_{i,j}}$ denotes the normalized distance between the $i,j$-th and $k,n$-th pixel values, $I_{i,j}$ denotes the $i,j$-th pixel value of the original image $I$, and $\sigma_{i,j}$ denotes the standard deviation of the image intensity computed locally for the affinity kernel. In addition, $G_{T}(M_{cam})[i,j]$ for $(i,j) \notin \mathcal{W}$ is set to zero.


Compared with existing methods (\emph{e.g.}, CRF \cite{krahenbuhl2011efficient}), PAMR effectively reduces the computational complexity by narrowing the affinity kernel computation $\alpha_{i,j,k,n}$ to regions of contiguous pixels $(k,n) \in \mathcal{N}(\cdot,\cdot)$ rather than all pixels $\{1:h,1:w\}$. 

\section{Additional analysis}
\label{section:analysis}

\subsection{Effect of joint consideration for RecurSeed and EdgePredictMix}
\label{ssec:effect}

To show the effects of RecurSeed (RS) and EdgePredictMix (EPM), we conduct ablation studies with and without RS and EPM individually on our single-stage network. Without RS and EPM, training is conducted only by minimizing the classification loss $L_{cls}$. 
Tab. \ref{tab:ablation} compares the network's CAM, SCG, and decoder results for up to 30 epochs. For the PASCAL VOC 2012 and MS COCO 2014 datasets, we observe an additional improvement in the mIoU of CAM and SCG results by approximately 3\%--5\% with each sequential application of RS and EPM, thereby demonstrating the individual necessity and effectiveness of RS and EPM. In addition, we verify the validity of the proposed decoder by showing that its results were consistently better than those of CAM and SCG by more than approximately 5\%.

\begin{table}[t]
  \caption{ 
    mIoUs ($\%$) of CAM, SCG, and the prediction from the segmentation branch (Decoder) on PASCAL VOC 2012 and MS COCO 2014 \emph{train} images. RS, RecurSeed; EPM, EdgePredictMix.  
  }
  \centering
  \begin{scriptsize} 
  \begin{tabular}{c | c c | c c c}
    \toprule
    Dataset & RS & EPM & CAM  & SCG  & Decoder \\
    \hline 
    \multirow{3}{*}{\begin{tabular}[c]{@{}c@{}}VOC\end{tabular}} &        &        & 47.4 & 58.0 & 17.4 \\
    & \checkmark &            & 57.9 & 65.9 & 70.7 \\
    \rowcolor{maroon!25}
    & \checkmark & \checkmark & \textbf{63.4} & \textbf{69.0} & \textbf{75.2} \\
    \hline 
    \multirow{3}{*}{\begin{tabular}[c]{@{}c@{}}COCO\end{tabular}} &       &        & 32.1 & 37.8 & 10.6 \\
    & \checkmark &            & 40.0 & 41.5 & 47.2 \\
    \rowcolor{maroon!25}
    & \checkmark & \checkmark & \textbf{42.3} & \textbf{43.2} & \textbf{50.3} \\
    \bottomrule
  \end{tabular}
  \label{tab:ablation}
  \end{scriptsize}
\end{table}

\subsection{Effect of high-order feature correlation in RecurSeed}
\label{ssec:effect_of_high_order}
In Tab. \ref{tab:scg}, we show the SCG results obtained by varying the layer combination in the $HSC$. For each combination, we display the mIoU of both cases, \emph{i.e.}, the first case in which the SCG result is generated from the raw CAM (\emph{i.e.}, $t = 1$) and the second case in which the SCG result is updated by the proposed RS (\emph{i.e.}, $t = 30$). In both cases, we observe that combining more layers in HSC sequentially improves the mIoU (\emph{i.e.}, from 49.7\% to 58.0\% for $t = 1$, and 67.5\% to 69.0\% for $t = 30$). From this observation, we use all layers from layer1 to layer5 for training the RS. In particular, the proposed RS and EPM consistently improve the performance under all combinations of layers (\emph{e.g.}, the performance improves by 11\%, from 58.0\% to 69.0\%, when combining all layers), validating the usefulness of our method.

\begin{table}
  \centering
  \caption{ 
    {Ablation study for each combination of the SCG. Our method boosts the overall performance of SCG applied in various layers. \checkmark indicates that SCG is applied. L5, L4, L3, L2, and L1 means layer5, layer4, layer3, layer2, and layer1, respectively.}
  }
  \vspace{-0.20cm}
  \begin{scriptsize} 
  \begin{tabular}{p{0.010\textwidth} p{0.010\textwidth} p{0.010\textwidth} p{0.010\textwidth} p{0.01\textwidth} | c c}
    \toprule
    L5         & L4         & L3         & L2         & L1     & $\tau$=1 (w.o. RS and EPM) & $\tau$=30 (w. RS and EPM) \\
    \hline 
    \checkmark &      &      &      &      & 49.7 & \textbf{67.5} \\
    \checkmark & \checkmark &      &      &      & 57.0 & \textbf{68.4} \\
    \checkmark & \checkmark & \checkmark &      &      & 57.7 & \textbf{68.8} \\
    \checkmark & \checkmark & \checkmark & \checkmark &      & 57.9 & \textbf{69.0} \\
    \rowcolor{maroon!25}
    \checkmark & \checkmark & \checkmark & \checkmark & \checkmark & 58.0 & \textbf{69.0} \\ 
    \bottomrule
  \end{tabular}
  \end{scriptsize}
  \label{tab:scg} 
\end{table}

In Tab. \ref{tab:scg2}, we add both FN and FP results into Tab. \ref{tab:scg}. Note that the closer the network layer is to the input, the more discriminative feature maps are extracted. Therefore, using high-order features (\emph{i.e.}, adding low layer maps L4-L1) reduces FP as it focuses on the target's discriminative area; however, it has the side effect of increasing FN in general. If RS is not used, this trend appears as shown in Tab. \ref{tab:scg}, but if RS is used, FP decreased ($0.232 \rightarrow 0.210$) without increasing FN ($0.161 \rightarrow 0.141$). As RS repeatedly gives diverse discriminative information to the network, the network can ideally accumulate all this information, thus compensating for the weakness of increasing FN. As such, RS does not increase FN but does not play a key role in reducing FN. However, it can be observed in Tab. \ref{tab:scg2} that FN is also significantly reduced ($0.148 \rightarrow 0.133$) when EPM is added to RS.  
The reason is that EPM can effectively reduce FN by letting the network better recognize a detailed region of each object. As a result, both FN and FP achieve the lowest values of 0.133 and 0.187, respectively, when RS and EPM are considered together.  

\begin{table}
  \centering
  \caption{ 
    Extended results of Tab. \ref{tab:scg} with mIoU, FP, and FN: w/o RS ($t = 1$), w/ RS ($t = 30$), L5 (layer5), and L4-L1 (layers 4+3+2+1).
  } 
  \vspace{-0.20cm}
  \begin{scriptsize} 
     \begin{tabular}{c c | c c | c c c}
        \toprule
        \multirow{2}{*}{\begin{tabular}[c]{@{}c@{}}RS\end{tabular}} & \multirow{2}{*}{\begin{tabular}[c]{@{}c@{}}EPM\end{tabular}} & Low       & High        & \multirow{2}{*}{\begin{tabular}[c]{@{}c@{}}mIoU\end{tabular}} & \multirow{2}{*}{\begin{tabular}[c]{@{}c@{}}FP\end{tabular}} & \multirow{2}{*}{\begin{tabular}[c]{@{}c@{}}FN\end{tabular}} \\
          & & L5        & L4-L1       &      &    &    \\
        \hline 
         &  & \checkmark &   & 49.7 & 0.358 & 0.159 \\
         &  & \checkmark & \checkmark & 58.0 & 0.268 (\textcolor{blue}{$\downarrow$ 0.090}) & 0.165 (\textcolor{red}{$\uparrow$ 0.006}) \\
        \hline
        \checkmark &  & \checkmark &   & 61.9 & 0.232 & 0.161 \\
        \checkmark &  & \checkmark & \checkmark & 65.9 & 0.210 (\textcolor{blue}{$\downarrow$ 0.022}) & 0.141 (\textcolor{blue}{$\downarrow$ 0.020}) \\
        \hline
        \checkmark & \checkmark & \checkmark &   & 67.5 & 0.188 & 0.148 \\
        \rowcolor{maroon!25}
        \checkmark & \checkmark & \checkmark & \checkmark & \textbf{69.0} & \textbf{0.187} (\textcolor{blue}{$\downarrow$ 0.001}) & \textbf{0.133} (\textcolor{blue}{$\downarrow$ 0.015}) \\
        \bottomrule
      \end{tabular}
      \label{tab:scg2}
    \end{scriptsize}
    \vspace{-0.20cm}
\end{table}

\subsection{Hyperparameters for Canny.}\label{ssec:canny_parameters} 
\begin{figure}[t]
  \centering
  \includegraphics[width=1.0\linewidth]{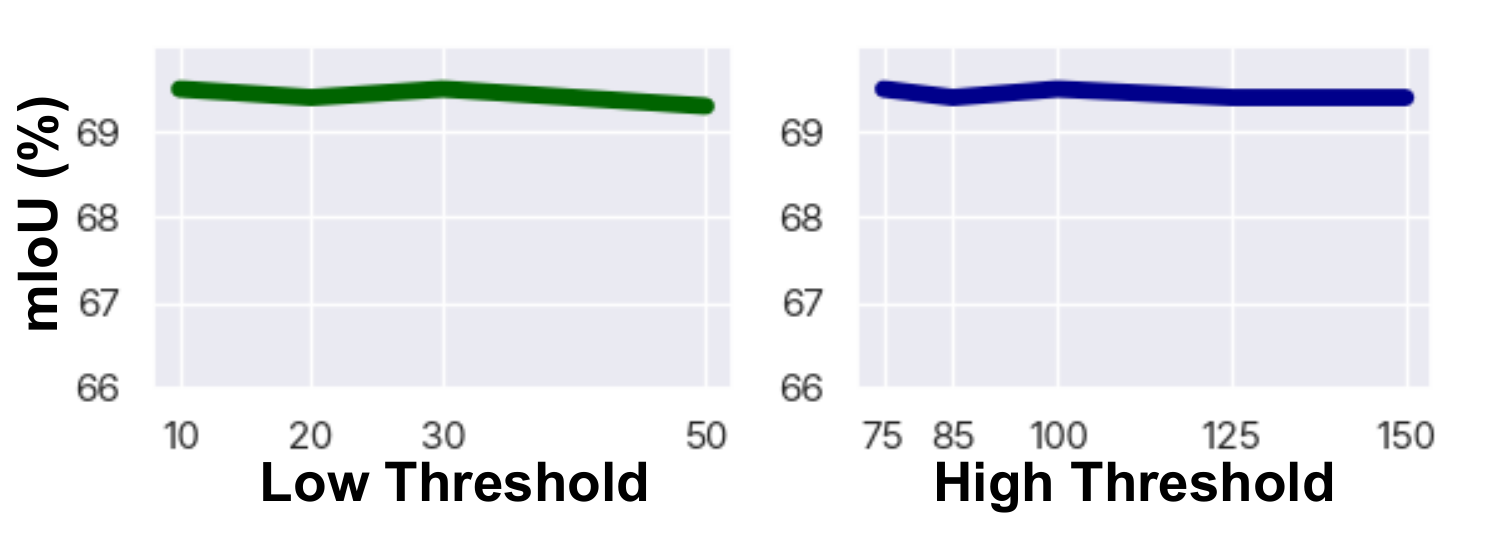}
  \vspace{-0.3cm}
  \caption{
      Sensitivity of low and high thresholds in Canny [Canny, 1986] on PASCAL VOC 2012 $val$ set. 
  }
  \label{fig:canny}
  \vspace{-0.40cm}
\end{figure}

To identify non-relevant and strong edges, we set low (10) and high (100) thresholds for Canny and 4-connectivity for Connected-component labeling (CCL) \cite{rosenfeld1966sequential}. We conduct Canny's parametric ablation study, as in Fig. \ref{fig:canny}, which shows our method is not sensitive to changing Canny's thresholds.

\section{Additional results}
\label{section:additional_results}

\subsection{Quantitative results.}\label{ssec:supple_quantitative} We show the per-class segmentation results from the both PASCAL VOC 2012 and MS COCO 2014 datasets in Tabs. \ref{tab:voc_val_detail}, \ref{tab:voc_test_detail}, and \ref{tab:coco_val_detail}. These quantitative results indicate that our method outperforms existing state-of-the-art methods in most class categories. 

\subsection{Qualitative results.}\label{ssec:supple_qualitative}  
\begin{figure*}[t]%
    \centering
    \includegraphics[width=1.0\linewidth,height=21cm]{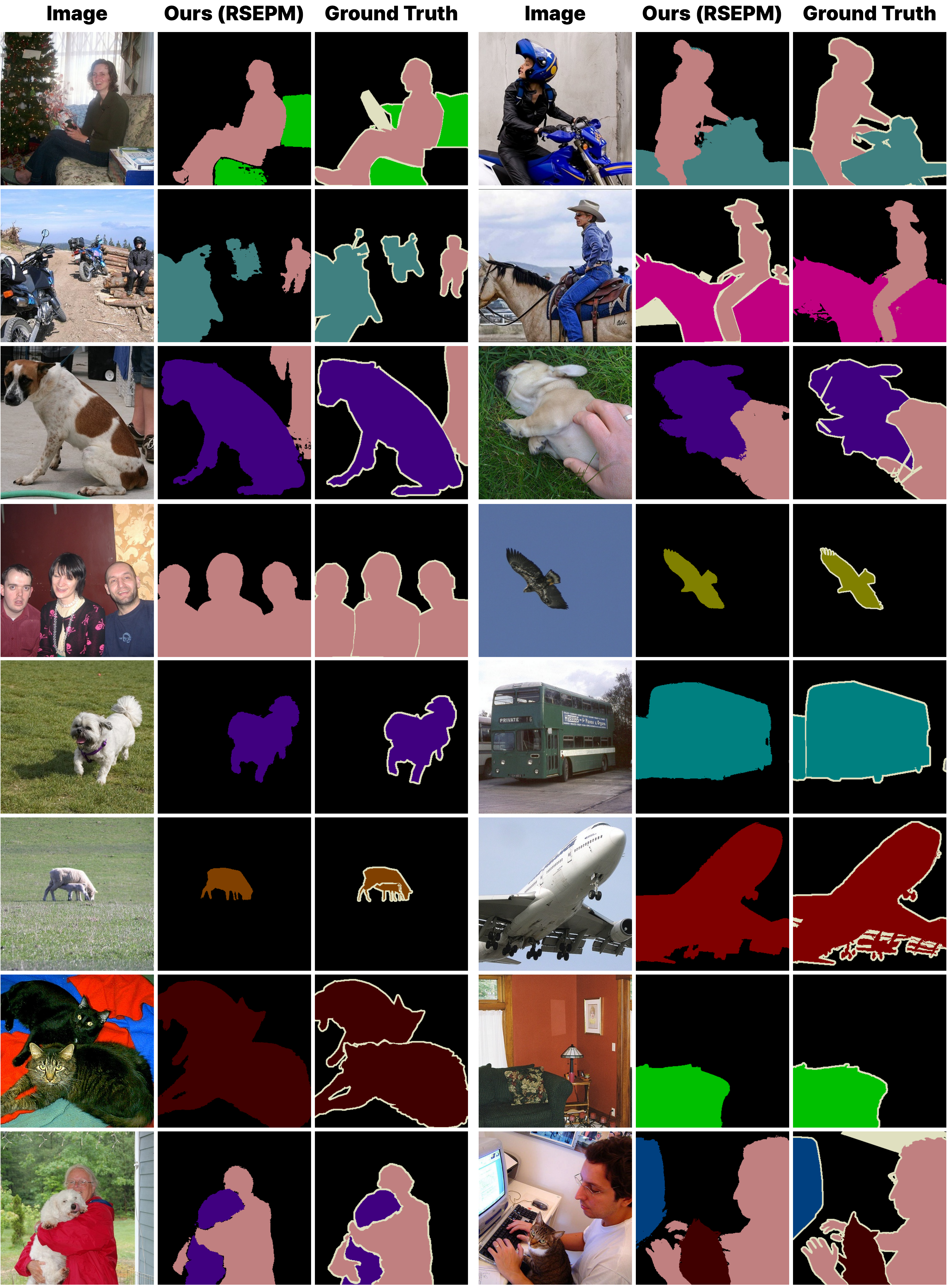}
    \caption[Short caption]{
        Qualitative segmentation results of PASCAL VOC 2012 \emph{val} set.
    }
    \label{fig:additional_voc_results}
\end{figure*}

\begin{figure*}[t]%
    \centering%
    \includegraphics[width=\linewidth]{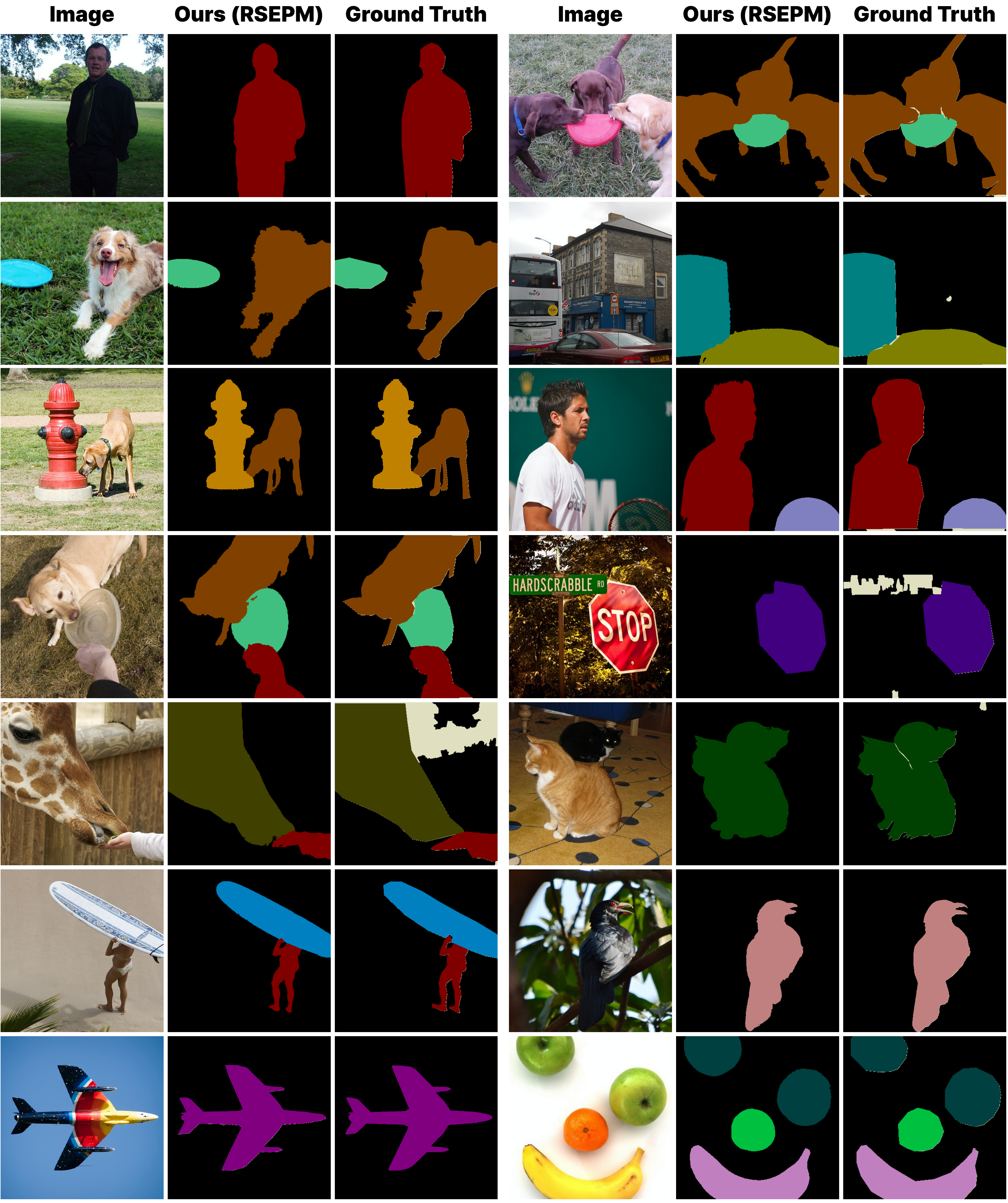}
    \caption{
        Qualitative segmentation results of MS COCO 2014 \emph{val} set.
    }
    \label{fig:additional_coco_results}
\end{figure*}

\begin{figure*}[t]%
    \centering%
    \includegraphics[width=1.0\linewidth]{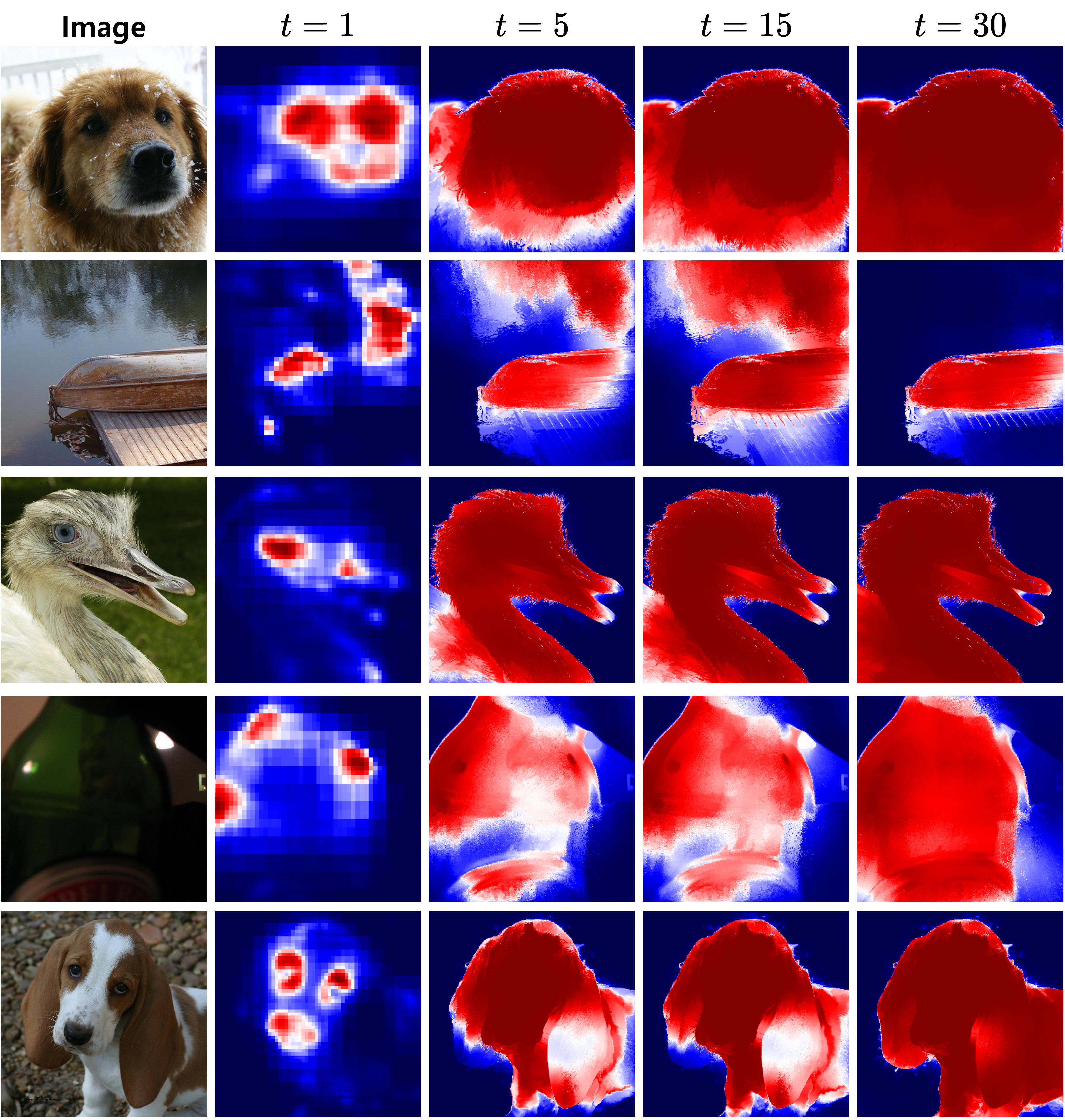}
    \caption{
        Visualization of attention maps with recursive improvements on the PASCAL VOC 2012 \emph{train} set.
    }
    \label{fig:additional_voc_att_results}
\end{figure*}

\begin{figure*}[t]%
    \centering%
    \includegraphics[width=1.0\linewidth]{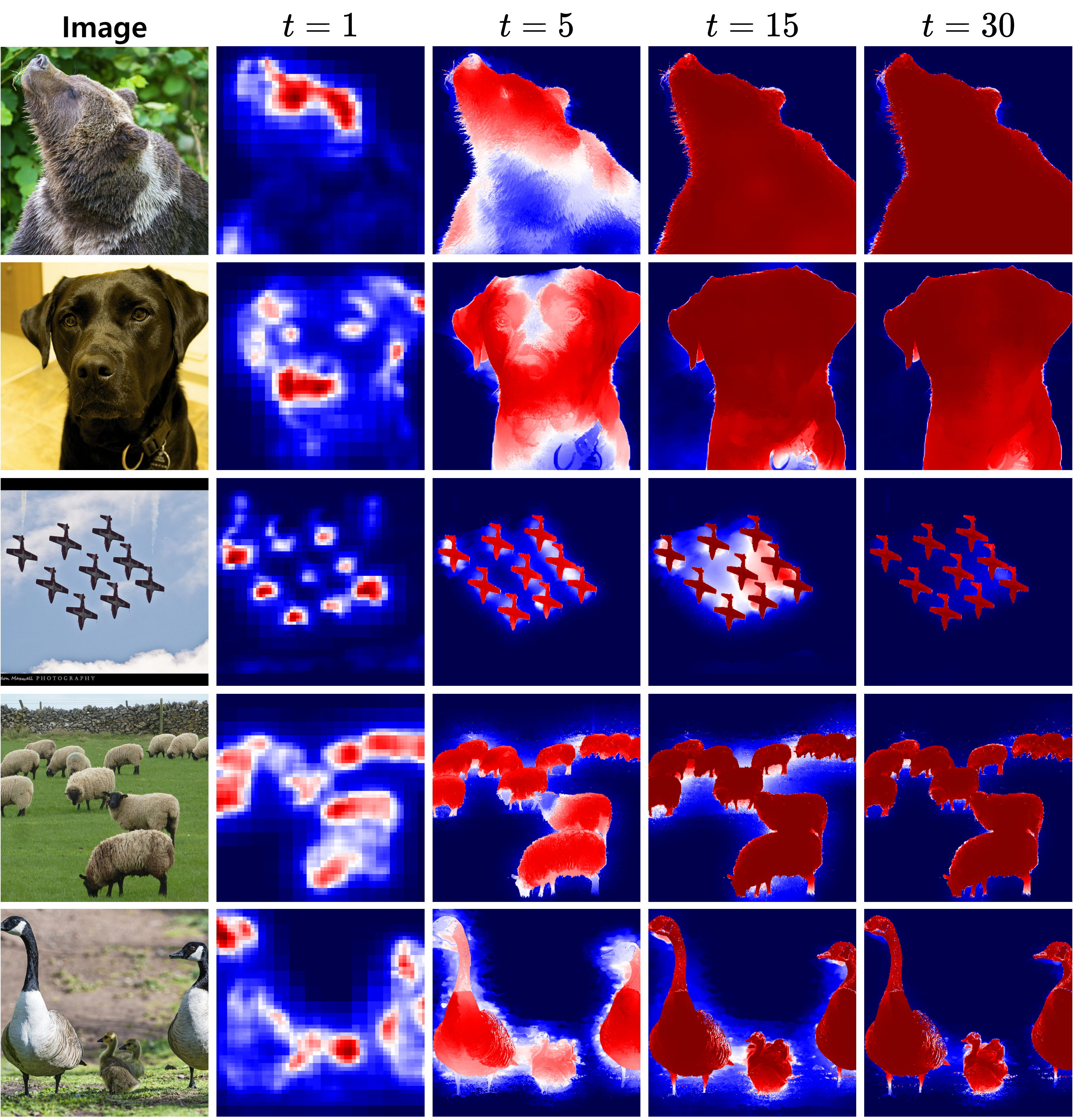}
    \caption{
        Visualization of attention maps with recursive improvements on the MS COCO 2014 \emph{train} set.
    }
    \label{fig:additional_coco_att_results}
\end{figure*}

We present additional examples of qualitative segmentation results produced by our method on both the PASCAL VOC 2012 and MS COCO 2014 $val$ sets in Figs. \ref{fig:additional_voc_results} and \ref{fig:additional_coco_results}, respectively. These results show that our method not only performs well for different complex scenes, small objects, or multiple instances but also can achieve a satisfactory segmentation performance for various challenging scenes. We also visualize more examples of attention maps for each step ($t = 1,5,15,30$) on the PASCAL VOC 2012 and MS COCO 2014 \emph{train} sets, as shown in Figs. \ref{fig:additional_voc_att_results} and \ref{fig:additional_coco_att_results}, respectively. The raw CAMs without RecurSeed (RS) (\emph{i.e.}, $t = 1$) only focus on the local discriminative parts for large-scale objects, such as the head and hands of people or wheels of the vehicles. However, when applying the proposed RS (\emph{i.e.}, $t = 5,15$), our single-stage results cover more object regions, including those less discriminative regions for large-scale objects, and also capture the exact boundaries of small-scale objects (\emph{i.e.}, decreasing the FP). Furthermore, with more training steps, including the proposed EdgePredictMix (EPM) (\emph{i.e.}, $t = 30$), our final results produce more accurate boundaries. This demonstrates the sequential improvements through the proposed RS and EPM.

\clearpage

\begin{table*}[t]
  \centering
  \caption{
    Per-class performance comparisons with WSSS methods in terms of IoUs (\%) on the PASCAL VOC 2012 \emph{val} set.
  }
  \begin{scriptsize}
  \begin{tabular}{
    p{0.17\textwidth} 
    p{0.015\textwidth} p{0.015\textwidth} p{0.015\textwidth} p{0.015\textwidth} p{0.015\textwidth} 
    p{0.015\textwidth} p{0.015\textwidth} p{0.015\textwidth} p{0.015\textwidth} p{0.015\textwidth} 
    p{0.015\textwidth} p{0.015\textwidth} p{0.015\textwidth} p{0.015\textwidth} p{0.015\textwidth} 
    p{0.015\textwidth} p{0.015\textwidth} p{0.015\textwidth} p{0.015\textwidth} p{0.015\textwidth} p{0.015\textwidth} 
    c }
    \toprule
    Method & \rotatebox[origin=l]{90}{bkg} & \rotatebox[origin=l]{90}{aero} & \rotatebox[origin=l]{90}{bike} & \rotatebox[origin=l]{90}{bird} & \rotatebox[origin=l]{90}{boat} & \rotatebox[origin=l]{90}{bottle} & \rotatebox[origin=l]{90}{bus} & \rotatebox[origin=l]{90}{car} & \rotatebox[origin=l]{90}{cat} & \rotatebox[origin=l]{90}{chair} & \rotatebox[origin=l]{90}{cow} & \rotatebox[origin=l]{90}{table} & \rotatebox[origin=l]{90}{dog} & \rotatebox[origin=l]{90}{horse} & \rotatebox[origin=l]{90}{mbk} & \rotatebox[origin=l]{90}{person} & \rotatebox[origin=l]{90}{plant} & \rotatebox[origin=l]{90}{sheep} & \rotatebox[origin=l]{90}{sofa} & \rotatebox[origin=l]{90}{train} & \rotatebox[origin=l]{90}{tv}  & mIoU  \\
    \hline \hline
    EM-Adapt \cite{papandreou2015weakly} & 67.2 & 29.2 & 17.6 & 28.6 & 22.2 & 29.6 & 47.0 & 44.0 & 44.2 & 14.6 & 35.1 & 24.9 & 41.0 & 34.8 & 41.6 & 32.1 & 24.8 & 37.4 & 24.0 & 38.1 & 31.6 & 33.8 \\
    MIL-LSE \cite{pinheiro2015image} & 79.6 & 50.2 & 21.6 & 40.9 & 34.9 & 40.5 & 45.9 & 51.5 & 60.6 & 12.6 & 51.2 & 11.6 & 56.8 & 52.9 & 44.8 & 42.7 & 31.2 & 55.4 & 21.5 & 38.8 & 36.9 & 42.0 \\
    SEC \cite{kolesnikov2016seed} & 82.4 & 62.9 & 26.4 & 61.6 & 27.6 & 38.1 & 66.6 & 62.7 & 75.2 & 22.1 & 53.5 & 28.3 & 65.8 & 57.8 & 62.3 & 52.5 & 32.5 & 62.6 & 32.1 & 45.4 & 45.3 & 50.7 \\
    TransferNet \cite{hong2016learning} & 85.3 & 68.5 & 26.4 & 69.8 & 36.7 & 49.1 & 68.4 & 55.8 & 77.3 & 6.2 & 75.2 & 14.3 & 69.8 & 71.5 & 61.1 & 31.9 & 25.5 & 74.6 & 33.8 & 49.6 & 43.7 & 52.1 \\
    CRF-RNN \cite{roy2017combining} & 85.8 & 65.2 & 29.4 & 63.8 & 31.2 & 37.2 & 69.6 & 64.3 & 76.2 & 21.4 & 56.3 & 29.8 & 68.2 & 60.6 & 66.2 & 55.8 & 30.8 & 66.1 & 34.9 & 48.8 & 47.1 & 52.8 \\
    WebCrawl \cite{hong2017weakly} & 87.0 & 69.3 & 32.2 & 70.2 & 31.2 & 58.4 & 73.6 & 68.5 & 76.5 & 26.8 & 63.8 & 29.1 & 73.5 & 69.5 & 66.5 & 70.4 & 46.8 & 72.1 & 27.3 & 57.4 & 50.2 & 58.1 \\
    DSRG \cite{huang2018weakly} & - & - & - & - & - & - & - & - & - & - & - & - & - & - & - & - & - & - & - & - & - & 61.4 \\
    PSA \cite{ahn2018learning} & 87.6 & 76.7 & 33.9 & 74.5 & 58.5 & 61.7 & 75.9 & 72.9 & 78.6 & 18.8 & 70.8 & 14.1 & 68.7 & 69.6 & 69.5 & 71.3 & 41.5 & 66.5 & 16.4 & 70.2 & 48.7 & 59.4 \\
    FickleNet \cite{lee2019ficklenet} & 89.5 & 76.6 & 32.6 & 74.6 & 51.5 & 71.1 & 83.4 & 74.4 & 83.6 & 24.1 & 73.4 & 47.4 & 78.2 & 74.0 & 68.8 & 73.2 & 47.8 & 79.9 & 37.0 & 57.3 & \textbf{64.6} & 64.9 \\
    SSDD \cite{shimoda2019self} & 89.0 & 62.5 & 28.9 & 83.7 & 52.9 & 59.5 & 77.6 & 73.7 & 87.0 & 34.0 & 83.7 & 47.6 & 84.1 & 77.0 & 73.9 & 69.6 & 29.8 & 84.0 & 43.2 & 68.0 & 53.4 & 64.9 \\
    MCIS \cite{sun2020mining} & - & - & - & - & - & - & - & - & - & - & - & - & - & - & - & - & - & - & - & - & - & 66.2 \\
    RRM \cite{zhang2020reliability} & 87.9 & 75.9 & 31.7 & 78.3 & 54.6 & 62.2 & 80.5 & 73.7 & 71.2 & 30.5 & 67.4 & 40.9 & 71.8 & 66.2 & 70.3 & 72.6 & 49.0 & 70.7 & 38.4 & 62.7 & 58.4 & 62.6 \\
    SSSS \cite{araslanov2020single} & 88.7 & 70.4 & 35.1 & 75.7 & 51.9 & 65.8 & 71.9 & 64.2 & 81.1 & 30.8 & 73.3 & 28.1 & 81.6 & 69.1 & 62.6 & 74.8 & 48.6 & 71.0 & 40.1 & 68.5 & 64.3 & 62.7 \\
    CONTA \cite{zhang2020causal} & - & - & - & - & - & - & - & - & - & - & - & - & - & - & - & - & - & - & - & - & - & 66.1 \\
    SEAM \cite{wang2020self} & 88.8 & 68.5 & 33.3 & 85.7 & 40.4 & 67.3 & 78.9 & 76.3 & 81.9 & 29.1 & 75.5 & 48.1 & 79.9 & 73.8 & 71.4 & 75.2 & 48.9 & 79.8 & 40.9 & 58.2 & 53.0 & 64.5 \\
    CIAN \cite{fan2020cian} & 88.2 & 79.5 & 32.6 & 75.7 & 56.8 & 72.1 & 85.3 & 72.9 & 81.7 & 27.6 & 73.3 & 39.8 & 76.4 & 77.0 & 74.9 & 66.8 & 46.6 & 81.0 & 29.1 & 60.4 & 53.3 & 64.3 \\
    NSRM \cite{yao2021non} & - & - & - & - & - & - & - & - & - & - & - & - & - & - & - & - & - & - & - & - & - & 68.3 \\
    EDAM \cite{wu2021embedded} & - & - & - & - & - & - & - & - & - & - & - & - & - & - & - & - & - & - & - & - & - & 70.9 \\
    AdvCAM \cite{lee2021anti} & 90.0 & 79.8 & 34.1 & 82.6 & 63.3 & 70.5 & 89.4 & 76.0 & 87.3 & 31.4 & 81.3 & 33.1 & 82.5 & 80.8 & 74.0 & 72.9 & 50.3 & 82.3 & 42.2 & \textbf{74.1} & 52.9 & 68.1 \\
    CSE \cite{kweon2021unlocking} & - & - & - & - & - & - & - & - & - & - & - & - & - & - & - & - & - & - & - & - & - & 68.4 \\
    DRS \cite{kim2021discriminative} & - & - & - & - & - & - & - & - & - & - & - & - & - & - & - & - & - & - & - & - & - & 70.4 \\
    CPN \cite{zhang2021complementary} & 89.9 & 75.0 & 32.9 & 87.8 & 60.9 & 69.4 & 87.7 & 79.4 & 88.9 & 28.0 & 80.9 & 34.8 & 83.4 & 79.6 & 74.6 & 66.9 & 56.4 & 82.6 & 44.9 & 73.1 & 45.7 & 67.8 \\
    RIB \cite{lee2021reducing} & 90.3 & 76.2 & 33.7 & 82.5 & \textbf{64.9} & 73.1 & 88.4 & 78.6 & 88.7 & 32.3 & 80.1 & 37.5 & 83.6 & 79.7 & 75.8 & 71.8 & 47.5 & 84.3 & 44.6 & 65.9 & 54.9 & 68.3 \\
    \rowcolor{maroon!25}
    Ours (single-stage, RS)& 89.7 & 80.0 & 36.1 & 87.7 & 40.1 & 65.2 & 82.6 & 75.3 & 88.6 & 30.1 & 74.4 & 48.9 & 82.9 & 79.0 & 75.0 & 80.9 & 46.0 & 75.7 & 47.1 & 52.1 & 58.3 & 66.5 \\
    \rowcolor{maroon!25}
    Ours (single-stage, RSEPM)& 91.3 & 85.7 & \textbf{40.0} & \textbf{88.1} & 52.7 & 67.3 & 85.9 & 80.1 & 89.2 & 32.4 & 78.5 & 48.7 & 83.9 & 81.2 & 77.3 & \textbf{84.7} & 52.2 & 83.3 & 46.7 & 51.2 & 59.4 & 69.5 \\
    \rowcolor{maroon!25}
    Ours (multi-stage, RS)& 91.7 & 85.0 & 32.5 & 87.5 & 48.3 & 79.4 & 91.7 & 83.0 & 92.5 & 38.2 & 88.2 & \textbf{60.8} & 89.7 & 86.2 & 79.9 & 83.3 & 56.0 & 85.3 & \textbf{57.2} & 55.5 & 56.3 & 72.8 \\
    \rowcolor{maroon!25}
    Ours (multi-stage, RSEPM)& \textbf{92.2} & \textbf{88.4} & 35.4 & 87.9 & 63.8 & \textbf{79.5} & \textbf{93.0} & \textbf{84.5} & \textbf{92.7} & \textbf{39.0} & \textbf{90.5} & 54.5 & \textbf{90.6} & \textbf{87.5} & \textbf{83.0} & 84.0 & \textbf{61.1} & \textbf{85.6} & 52.1 & 56.2 & 60.2 & \textbf{74.4} \\
    \hline
    \bottomrule
  \end{tabular}
  \end{scriptsize}
  \label{tab:voc_val_detail}
\end{table*}

\begin{table*}[t]
  \centering
  \caption{
    Per-class performance comparisons with WSSS methods in terms of IoUs (\%) on the PASCAL VOC 2012 \emph{test} set.
  }
  \begin{scriptsize}
  \begin{tabular}{
    p{0.17\textwidth} 
    p{0.015\textwidth} p{0.015\textwidth} p{0.015\textwidth} p{0.015\textwidth} p{0.015\textwidth} 
    p{0.015\textwidth} p{0.015\textwidth} p{0.015\textwidth} p{0.015\textwidth} p{0.015\textwidth} 
    p{0.015\textwidth} p{0.015\textwidth} p{0.015\textwidth} p{0.015\textwidth} p{0.015\textwidth} 
    p{0.015\textwidth} p{0.015\textwidth} p{0.015\textwidth} p{0.015\textwidth} p{0.015\textwidth} p{0.015\textwidth} 
    c }
    \toprule
    Method & \rotatebox[origin=l]{90}{bkg} & \rotatebox[origin=l]{90}{aero} & \rotatebox[origin=l]{90}{bike} & \rotatebox[origin=l]{90}{bird} & \rotatebox[origin=l]{90}{boat} & \rotatebox[origin=l]{90}{bottle} & \rotatebox[origin=l]{90}{bus} & \rotatebox[origin=l]{90}{car} & \rotatebox[origin=l]{90}{cat} & \rotatebox[origin=l]{90}{chair} & \rotatebox[origin=l]{90}{cow} & \rotatebox[origin=l]{90}{table} & \rotatebox[origin=l]{90}{dog} & \rotatebox[origin=l]{90}{horse} & \rotatebox[origin=l]{90}{mbk} & \rotatebox[origin=l]{90}{person} & \rotatebox[origin=l]{90}{plant} & \rotatebox[origin=l]{90}{sheep} & \rotatebox[origin=l]{90}{sofa} & \rotatebox[origin=l]{90}{train} & \rotatebox[origin=l]{90}{tv}  & mIoU  \\
    \hline \hline
    EM-Adapt \cite{papandreou2015weakly} & 76.3 & 37.1 & 21.9 & 41.6 & 26.1 & 38.5 & 50.8 & 44.9 & 48.9 & 16.7 & 40.8 & 29.4 & 47.1 & 45.8 & 54.8 & 28.2 & 30.0 & 44.0 & 29.2 & 34.3 & 46.0 & 39.6 \\
    MIL-LSE \cite{pinheiro2015image} & 78.7 & 48.0 & 21.2 & 31.1 & 28.4 & 35.1 & 51.4 & 55.5 & 52.8 & 7.8 & 56.2 & 19.9 & 53.8 & 50.3 & 40.0 & 38.6 & 27.8 & 51.8 & 24.7 & 33.3 & 46.3 & 40.6 \\
    SEC \cite{kolesnikov2016seed} & 83.5 & 56.4 & 28.5 & 64.1 & 23.6 & 46.5 & 70.6 & 58.5 & 71.3 & 23.2 & 54.0 & 28.0 & 68.1 & 62.1 & 70.0 & 55.0 & 38.4 & 58.0 & 39.9 & 38.4 & 48.3 & 51.7 \\
    TransferNet \cite{hong2016learning} & 85.7 & 70.1 & 27.8 & 73.7 & 37.3 & 44.8 & 71.4 & 53.8 & 73.0 & 6.7 & 62.9 & 12.4 & 68.4 & 73.7 & 65.9 & 27.9 & 23.5 & 72.3 & 38.9 & 45.9 & 39.2 & 51.2 \\
    CRF-RNN \cite{roy2017combining} & 85.7 & 58.8 & 30.5 & 67.6 & 24.7 & 44.7 & 74.8 & 61.8 & 73.7 & 22.9 & 57.4 & 27.5 & 71.3 & 64.8 & 72.4 & 57.3 & 37.3 & 60.4 & 42.8 & 42.2 & 50.6 & 53.7 \\
    WebCrawl \cite{hong2017weakly} & 87.2 & 63.9 & 32.8 & 72.4 & 26.7 & 64.0 & 72.1 & 70.5 & 77.8 & 23.9 & 63.6 & 32.1 & 77.2 & 75.3 & 76.2 & 71.5 & 45.0 & 68.8 & 35.5 & 46.2 & 49.3 & 58.7 \\
    DSRG \cite{huang2018weakly} & - & - & - & - & - & - & - & - & - & - & - & - & - & - & - & - & - & - & - & - & - & 63.2 \\
    PSA \cite{ahn2018learning} & 89.1 & 70.6 & 31.6 & 77.2 & 42.2 & 68.9 & 79.1 & 66.5 & 74.9 & 29.6 & 68.7 & 56.1 & 82.1 & 64.8 & 78.6 & 73.5 & 50.8 & 70.7 & 47.7 & 63.9 & 51.1 & 63.7 \\
    FickleNet \cite{lee2019ficklenet} & 90.3 & 77.0 & 35.2 & 76.0 & 54.2 & 64.3 & 76.6 & 76.1 & 80.2 & 25.7 & 68.6 & 50.2 & 74.6 & 71.8 & 78.3 & 69.5 & 53.8 & 76.5 & 41.8 & \textbf{70.0} & 54.2 & 65.0 \\
    SSDD \cite{shimoda2019self} & 89.5 & 71.8 & 31.4 & 79.3 & 47.3 & 64.2 & 79.9 & 74.6 & 84.9 & 30.8 & 73.5 & 58.2 & 82.7 & 73.4 & 76.4 & 69.9 & 37.4 & 80.5 & 54.5 & 65.7 & 50.3 & 65.5 \\
    MCIS \cite{sun2020mining} & - & - & - & - & - & - & - & - & - & - & - & - & - & - & - & - & - & - & - & - & - & 66.9 \\
    RRM \cite{zhang2020reliability} & 87.8 & 77.5 & 30.8 & 71.7 & 36.0 & 64.2 & 75.3 & 70.4 & 81.7 & 29.3 & 70.4 & 52.0 & 78.6 & 73.8 & 74.4 & 72.1 & 54.2 & 75.2 & 50.6 & 42.0 & 52.5 & 62.9 \\
    SSSS \cite{araslanov2020single} & 88.7 & 70.4 & 35.1 & 75.7 & 51.9 & 65.8 & 71.9 & 64.2 & 81.1 & 30.8 & 73.3 & 28.1 & 81.6 & 69.1 & 62.6 & 74.8 & 48.6 & 71.0 & 40.1 & 68.5 & \textbf{64.3} & 62.7 \\
    CONTA \cite{zhang2020causal} & - & - & - & - & - & - & - & - & - & - & - & - & - & - & - & - & - & - & - & - & - & 66.7 \\
    SEAM \cite{wang2020self} & 88.8 & 68.5 & 33.3 & 85.7 & 40.4 & 67.3 & 78.9 & 76.3 & 81.9 & 29.1 & 75.5 & 48.1 & 79.9 & 73.8 & 71.4 & 75.2 & 48.9 & 79.8 & 40.9 & 58.2 & 53.0 & 64.5 \\
    NSRM \cite{yao2021non} & - & - & - & - & - & - & - & - & - & - & - & - & - & - & - & - & - & - & - & - & - & 68.5 \\
    EDAM \cite{wu2021embedded} & - & - & - & - & - & - & - & - & - & - & - & - & - & - & - & - & - & - & - & - & - & 70.6 \\
    AdvCAM \cite{lee2021anti} & 90.1 & 81.2 & 33.6 & 80.4 & 52.4 & 66.6 & 87.1 & 80.5 & 87.2 & 28.9 & 80.1 & 38.5 & 84.0 & 83.0 & 79.5 & 71.9 & 47.5 & 80.8 & \textbf{59.1} & 65.4 & 49.7 & 68.0 \\
    CSE \cite{kweon2021unlocking} & - & - & - & - & - & - & - & - & - & - & - & - & - & - & - & - & - & - & - & - & - & 68.2 \\
    DRS \cite{kim2021discriminative} & - & - & - & - & - & - & - & - & - & - & - & - & - & - & - & - & - & - & - & - & - & 70.7 \\
    CPN \cite{zhang2021complementary} & 90.4 & 79.8 & 32.9 & 85.7 & 52.8 & 66.3 & 87.2 & 81.3 & 87.6 & 28.2 & 79.7 & 50.1 & 82.9 & 80.4 & 78.8 & 70.6 & 51.1 & 83.4 & 55.4 & 68.5 & 44.6 & 68.5 \\
    RIB \cite{lee2021reducing} & 90.4 & 80.5 & 32.8 & 84.9 & 59.4 & 69.3 & 87.2 & 83.5 & 88.3 & 31.1 & 80.4 & 44.0 & 84.4 & 82.3 & 80.9 & 70.7 & 43.5 & 84.9 & 55.9 & 59.0 & 47.3 & 68.6 \\
    \rowcolor{maroon!25}
    Ours (single-stage, RS)& 89.8 & 83.8 & 33.4 & 87.5 & 39.8 & 67.1 & 85.2 & 78.5 & 91.2 & 29.3 & 77.9 & 54.1 & 84.3 & 81.7 & 78.5 & 78.9 & 53.4 & 78.4 & 53.9 & 42.2 & 56.9 & 67.9 \\
    \rowcolor{maroon!25}
    Ours (single-stage, RSEPM)& 91.2 & 86.8 & \textbf{37.3} & 80.6 & 52.1 & 71.3 & 87.8 & 81.3 & 90.9 & 32.2 & 80.4 & 54.5 & 86.4 & 88.6 & \textbf{83.6} & 80.8 & 59.0 & 82.8 & 54.2 & 42.6 & 59.1 & 70.6 \\
    \rowcolor{maroon!25}
    Ours (multi-stage, RS)& 91.4 & 89.5 & 36.2 & \textbf{88.7} & 46.2 & 69.1 & 93.1 & \textbf{85.7} & \textbf{91.8} & 34.0 & 86.7 & \textbf{66.3} & \textbf{89.0} & 88.8 & 82.9 & \textbf{81.3} & 57.3 & 89.0 & 57.5 & 46.9 & 57.9 & 72.8 \\
    \rowcolor{maroon!25}
    Ours (multi-stage, RSEPM)& \textbf{91.9} & \textbf{89.7} & 37.3 & 88.0 & \textbf{62.5} & \textbf{72.1} & \textbf{93.5} & 85.6 & 90.2 & \textbf{36.3} & \textbf{88.3} & 62.5 & 86.3 & \textbf{89.1} & 82.9 & 81.2 & \textbf{59.7} & \textbf{89.2} & 56.2 & 44.5 & 59.4 & \textbf{73.6} \\
    \hline
    \bottomrule
  \end{tabular}
  \end{scriptsize}
  \label{tab:voc_test_detail}
\end{table*}

\clearpage

\begin{table*}[t]
  \centering
  \caption{
    Per-class performance comparisons with WSSS methods in terms of IoUs (\%) on the MS COCO 2014 \emph{val} set.
  }
  \begin{tabular}{
    p{0.125\textwidth} c c c c | p{0.10\textwidth} c c c c}
    \toprule
    Class & SEC & DSRG & Ours (RS) & Ours (RSEPM) & Class &  SEC & DSRG & Ours (RS) & Ours (RSEPM) \\
    \hline \hline
    background & 74.3 & 80.6 & 82.6 & \textbf{83.6} & wine glass & 22.3 & 24.0 & 37.5 & \textbf{39.8} \\
    person & 43.6 & - & 74.8 & \textbf{74.9} & cup & 17.9 & 20.4 & \textbf{39.1} & 38.9 \\
    bicycle & 24.2 & 30.4 & 53.1 & \textbf{55.0} & fork & 1.8 & 0.0 & \textbf{19.9} & 4.9 \\
    car & 15.9 & 22.1 & 48.6 & \textbf{50.1} & knife & 1.4 & 5.0 & \textbf{19.9} & 9.0 \\
    motorcycle & 52.1 & 54.2 & 72.4 & \textbf{72.9} & spoon & 0.6 & 0.5 & \textbf{5.5} & 1.1 \\
    airplane & 36.6 & 45.2 & 73.4 & \textbf{76.5} & bowl & 12.5 & 18.8 & \textbf{26.8} & 11.3 \\
    bus & 37.7 & 38.7 & 71.2 & \textbf{72.5} & banana & 43.6 & 46.4 & 66.4 & \textbf{67.0} \\
    train & 30.1 & 33.2 & 44.8 & \textbf{47.4} & apple & 23.6 & 24.3 & 43.0 & \textbf{49.2} \\
    truck & 24.1 & 25.9 & \textbf{46.5} & \textbf{46.5} & sandwich & 22.8 & 24.5 & \textbf{39.7} & 33.7 \\
    boat & 17.3 & 20.6 & 32.1 & \textbf{44.1} & orange & 44.3 & 41.2 & 59.8 & \textbf{62.3} \\
    traffic light & 16.7 & 16.1 & 23.6 & \textbf{60.8} & broccoli & 36.8 & 35.7 & 46.5 & \textbf{50.4} \\
    fire hydrant & 55.9 & 60.4 & 79.0 & \textbf{80.3} & carrot & 6.7 & 15.3 & \textbf{35.1} & 35.0 \\
    stop sign & 48.4 & 51.0 & 79.0 & \textbf{84.1} & hot dog & 31.2 & 24.9 & \textbf{49.0} & 48.3 \\
    parking meter & 25.2 & 26.3 & 72.2 & \textbf{77.8} & pizza & 50.9 & 56.2 & \textbf{69.9} & 68.6 \\
    bench & 16.4 & 22.3 & 40.3 & \textbf{41.2} & donut & 32.8 & 34.2 & \textbf{62.6} & 62.3 \\
    bird & 34.7 & 41.5 & \textbf{65.2} & 62.6 & cake & 12.0 & 6.9 & \textbf{50.7} & 48.3 \\
    cat & 57.2 & 62.2 & \textbf{79.2} & \textbf{79.2} & chair & 7.8 & 9.7 & 26.9 & \textbf{28.9} \\
    dog & 45.2 & 55.6 & \textbf{73.4} & 73.3 & couch & 5.6 & 17.7 & \textbf{47.0} & 44.9 \\
    horse & 34.4 & 42.3 & 74.4 & \textbf{76.1} & potted plant & 6.2 & 14.3 & \textbf{20.3} & 16.9 \\
    sheep & 40.3 & 47.1 & 76.4 & \textbf{80.0} & bed & 23.4 & 32.4 & \textbf{54.8} & 53.6 \\
    cow & 41.4 & 49.3 & 78.4 & \textbf{79.3} & dining table & 0.0 & 3.8 & \textbf{31.4} & 24.6 \\
    elephant & 62.9 & 67.1 & 84.7 & \textbf{85.6} & toilet & 38.5 & 43.6 & \textbf{71.1} & \textbf{71.1} \\
    bear & 59.1 & 62.6 & \textbf{84.9} & 82.9 & tv & 19.2 & 25.3 & 49.5 & \textbf{49.9} \\
    zebra & 59.8 & 63.2 & 85.2 & \textbf{87.0} & laptop & 20.1 & 21.1 & \textbf{57.0} & 56.6 \\
    giraffe & 48.8 & 54.3 & 79.8 & \textbf{82.2} & mouse & 3.5 & 0.9 & 7.9 & \textbf{17.4} \\
    backpack & 0.3 & 0.2 & \textbf{21.6} & 9.4 & remote & 17.5 & 20.6 & 50.3 & \textbf{54.8} \\
    umbrella & 26.0 & 35.3 & 71.1 & \textbf{73.4} & keyboard & 12.5 & 12.3 & \textbf{51.1} & 48.8 \\
    handbag & 0.5 & 0.7 & \textbf{9.3} & 4.6 & cell phone & 32.1 & 33.0 & \textbf{61.2} & 60.8 \\
    tie & 6.5 & 7.0 & \textbf{18.0} & 17.2 & microwave & 8.2 & 11.2 & \textbf{47.7} & 43.6 \\
    suitcase & 16.7 & 23.4 & \textbf{54.6} & 53.9 & oven & 13.7 & 12.4 & \textbf{42.2} & 38.0 \\
    frisbee & 12.3 & 13.0 & 55.1 & \textbf{57.7} & toaster & 0.0 & 0.0 & \textbf{0.2} & 0.0 \\
    skis & 1.6 & 1.5 & \textbf{9.0} & 8.2 & sink & 10.8 & 17.8 & \textbf{38.8} & 36.9 \\
    snowboard & 5.3 & 16.3 & \textbf{28.6} & 24.7 & refrigerator & 4.0 & 15.5 & \textbf{59.4} & 51.8 \\
    sports ball & 7.9 & 9.8 & 24.5 & \textbf{41.6} & book & 0.4 & 12.3 & \textbf{28.6} & 27.3 \\
    kite & 9.1 & 17.4 & 31.2 & \textbf{62.6} & clock & 17.8 & 20.7 & \textbf{27.8} & 23.3 \\
    baseball bat & 1.0 & \textbf{4.8} & 1.4 & 1.5 & vase & 18.4 & 23.9 & \textbf{27.1} & 26.0 \\
    baseball glove & 0.6 & \textbf{1.2} & 1.1 & 0.4 & scissors & 16.5 & 17.3 & 46.6 & \textbf{47.1} \\
    skateboard & 7.1 & 14.4 & 30.4 & \textbf{34.8} & teddy bear & 47.0 & 46.3 & \textbf{69.2} & 68.8 \\
    surfboard & 7.7 & 13.5 & 11.1 & \textbf{17.0} & hair drier & 0.0 & 0.0 & 0.0 & 0.0 \\
    tennis racket & 9.1 & 6.8 & \textbf{14.8} & 9.0 & toothbrush & 2.8 & 2.0 & \textbf{23.3} & 19.7 \\\cline{6-10}
    bottle & 13.2 & 22.3 & \textbf{41.3} & 38.1 & \textbf{mIoU} & 22.4 & 26.0 & 45.8 & \textbf{46.4} \\
    \bottomrule
  \end{tabular}
  \label{tab:coco_val_detail}
\end{table*}

\clearpage

\clearpage

\clearpage

\clearpage

\clearpage

{
    \small
    \bibliographystyle{ieeenat_fullname}
    \bibliography{main}

\begin{thebibliography}{60}
\providecommand{\natexlab}[1]{#1}
\providecommand{\url}[1]{\texttt{#1}}
\expandafter\ifx\csname urlstyle\endcsname\relax
  \providecommand{\doi}[1]{doi: #1}\else
  \providecommand{\doi}{doi: \begingroup \urlstyle{rm}\Url}\fi

\bibitem[Ahn and Kwak(2018)]{ahn2018learning}
Jiwoon Ahn and Suha Kwak.
\newblock Learning pixel-level semantic affinity with image-level supervision for weakly supervised semantic segmentation.
\newblock In \emph{IEEE CVPR}, pages 4981--4990, 2018.

\bibitem[Ahn et~al.(2019)Ahn, Cho, and Kwak]{ahn2019weakly}
Jiwoon Ahn, Sunghyun Cho, and Suha Kwak.
\newblock Weakly supervised learning of instance segmentation with inter-pixel relations.
\newblock In \emph{IEEE CVPR}, pages 2209--2218, 2019.

\bibitem[Araslanov and Roth(2020)]{araslanov2020single}
Nikita Araslanov and Stefan Roth.
\newblock Single-stage semantic segmentation from image labels.
\newblock In \emph{IEEE CVPR}, pages 4253--4262, 2020.

\bibitem[Bearman et~al.(2016)Bearman, Russakovsky, Ferrari, and Fei-Fei]{bearman2016s}
Amy Bearman, Olga Russakovsky, Vittorio Ferrari, and Li Fei-Fei.
\newblock What’s the point: {S}emantic segmentation with point supervision.
\newblock In \emph{ECCV}, pages 549--565. Springer, 2016.

\bibitem[Canny(1986)]{canny1986computational}
John Canny.
\newblock A computational approach to edge detection.
\newblock \emph{IEEE TPAMI}, pages 679--698, 1986.

\bibitem[Chen et~al.(2018)Chen, Zhu, Papandreou, Schroff, and Adam]{chen2018encoder}
Liang-Chieh Chen, Yukun Zhu, George Papandreou, Florian Schroff, and Hartwig Adam.
\newblock Encoder-decoder with atrous separable convolution for semantic image segmentation.
\newblock In \emph{ECCV}, pages 801--818, 2018.

\bibitem[Chen et~al.(2022{\natexlab{a}})Chen, Yang, Lai, and Xie]{chen2022self}
Qi Chen, Lingxiao Yang, Jian-Huang Lai, and Xiaohua Xie.
\newblock Self-supervised image-specific prototype exploration for weakly supervised semantic segmentation.
\newblock In \emph{IEEE CVPR}, pages 4288--4298, 2022{\natexlab{a}}.

\bibitem[Chen et~al.(2022{\natexlab{b}})Chen, Wang, Wu, Hua, Zhang, and Sun]{chen2022class}
Zhaozheng Chen, Tan Wang, Xiongwei Wu, Xian-Sheng Hua, Hanwang Zhang, and Qianru Sun.
\newblock Class re-activation maps for weakly-supervised semantic segmentation.
\newblock In \emph{IEEE CVPR}, pages 969--978, 2022{\natexlab{b}}.

\bibitem[Cheng et~al.(2023)Cheng, Qiao, Li, Li, Wei, Ji, Yuan, Liu, and Chen]{cheng2023out}
Zesen Cheng, Pengchong Qiao, Kehan Li, Siheng Li, Pengxu Wei, Xiangyang Ji, Li Yuan, Chang Liu, and Jie Chen.
\newblock Out-of-candidate rectification for weakly supervised semantic segmentation.
\newblock In \emph{IEEE CVPR}, pages 23673--23684, 2023.

\bibitem[Du et~al.(2022)Du, Fu, Liu, and Wang]{du2022weakly}
Ye Du, Zehua Fu, Qingjie Liu, and Yunhong Wang.
\newblock Weakly supervised semantic segmentation by pixel-to-prototype contrast.
\newblock In \emph{IEEE CVPR}, pages 4320--4329, 2022.

\bibitem[Everingham et~al.(2010)Everingham, Van~Gool, Williams, Winn, and Zisserman]{everingham2010pascal}
Mark Everingham, Luc Van~Gool, Christopher~KI Williams, John Winn, and Andrew Zisserman.
\newblock The pascal visual object classes ({VOC}) challenge.
\newblock \emph{IJCV}, 88\penalty0 (2):\penalty0 303--338, 2010.

\bibitem[Fan et~al.(2020)Fan, Zhang, Tan, Song, and Xiao]{fan2020cian}
Junsong Fan, Zhaoxiang Zhang, Tieniu Tan, Chunfeng Song, and Jun Xiao.
\newblock {C}ian: {C}ross-image affinity net for weakly supervised semantic segmentation.
\newblock In \emph{AAAI}, pages 10762--10769, 2020.

\bibitem[He et~al.(2016)He, Zhang, Ren, and Sun]{he2016deep}
Kaiming He, Xiangyu Zhang, Shaoqing Ren, and Jian Sun.
\newblock Deep residual learning for image recognition.
\newblock In \emph{IEEE CVPR}, pages 770--778, 2016.

\bibitem[Hong et~al.(2016)Hong, Oh, Lee, and Han]{hong2016learning}
Seunghoon Hong, Junhyuk Oh, Honglak Lee, and Bohyung Han.
\newblock {L}earning transferrable knowledge for semantic segmentation with deep convolutional neural network.
\newblock In \emph{IEEE CVPR}, pages 3204--3212, 2016.

\bibitem[Hong et~al.(2017)Hong, Yeo, Kwak, Lee, and Han]{hong2017weakly}
Seunghoon Hong, Donghun Yeo, Suha Kwak, Honglak Lee, and Bohyung Han.
\newblock {W}eakly supervised semantic segmentation using web-crawled videos.
\newblock In \emph{IEEE CVPR}, pages 7322--7330, 2017.

\bibitem[Huang et~al.(2018)Huang, Wang, Wang, Liu, and Wang]{huang2018weakly}
Zilong Huang, Xinggang Wang, Jiasi Wang, Wenyu Liu, and Jingdong Wang.
\newblock Weakly-supervised semantic segmentation network with deep seeded region growing.
\newblock In \emph{IEEE CVPR}, pages 7014--7023, 2018.

\bibitem[Jiang et~al.(2022)Jiang, Yang, Hou, and Wei]{jiang2022l2g}
Peng-Tao Jiang, Yuqi Yang, Qibin Hou, and Yunchao Wei.
\newblock {L2G}: A simple local-to-global knowledge transfer framework for weakly supervised semantic segmentation.
\newblock In \emph{IEEE CVPR}, pages 16886--16896, 2022.

\bibitem[Jo and Yu(2021)]{jo2021puzzle}
Sanghyun Jo and In-Jae Yu.
\newblock {Puzzle-CAM}: Improved localization via matching partial and full features.
\newblock In \emph{IEEE ICIP}, pages 639--643. IEEE, 2021.

\bibitem[Kim et~al.(2021)Kim, Han, and Kim]{kim2021discriminative}
Beomyoung Kim, Sangeun Han, and Junmo Kim.
\newblock Discriminative region suppression for weakly-supervised semantic segmentation.
\newblock In \emph{AAAI}, pages 1754--1761, 2021.

\bibitem[Kolesnikov and Lampert(2016)]{kolesnikov2016seed}
Alexander Kolesnikov and Christoph~H Lampert.
\newblock Seed, expand and constrain: {T}hree principles for weakly-supervised image segmentation.
\newblock In \emph{ECCV}, pages 695--711. Springer, 2016.

\bibitem[Kr{\"a}henb{\"u}hl and Koltun(2011)]{krahenbuhl2011efficient}
Philipp Kr{\"a}henb{\"u}hl and Vladlen Koltun.
\newblock {E}fficient inference in fully connected {CRF}s with gaussian edge potentials.
\newblock \emph{NeurlPS}, 24:\penalty0 109--117, 2011.

\bibitem[Kweon et~al.(2021)Kweon, Yoon, Kim, Park, and Yoon]{kweon2021unlocking}
Hyeokjun Kweon, Sung-Hoon Yoon, Hyeonseong Kim, Daehee Park, and Kuk-Jin Yoon.
\newblock Unlocking the potential of ordinary classifier: {C}lass-specific adversarial erasing framework for weakly supervised semantic segmentation.
\newblock In \emph{IEEE ICCV}, pages 6994--7003, 2021.

\bibitem[Kweon et~al.(2023)Kweon, Yoon, and Yoon]{kweon2023weakly}
Hyeokjun Kweon, Sung-Hoon Yoon, and Kuk-Jin Yoon.
\newblock Weakly supervised semantic segmentation via adversarial learning of classifier and reconstructor.
\newblock In \emph{IEEE CVPR}, pages 11329--11339, 2023.

\bibitem[Lee et~al.(2019)Lee, Kim, Lee, Lee, and Yoon]{lee2019ficklenet}
Jungbeom Lee, Eunji Kim, Sungmin Lee, Jangho Lee, and Sungroh Yoon.
\newblock {FickleNet}: {W}eakly and semi-supervised semantic image segmentation using stochastic inference.
\newblock In \emph{IEEE CVPR}, pages 5267--5276, 2019.

\bibitem[Lee et~al.(2021{\natexlab{a}})Lee, Choi, Mok, and Yoon]{lee2021reducing}
Jungbeom Lee, Jooyoung Choi, Jisoo Mok, and Sungroh Yoon.
\newblock Reducing information bottleneck for weakly supervised semantic segmentation.
\newblock \emph{NeurlPS}, 34, 2021{\natexlab{a}}.

\bibitem[Lee et~al.(2021{\natexlab{b}})Lee, Kim, and Yoon]{lee2021anti}
Jungbeom Lee, Eunji Kim, and Sungroh Yoon.
\newblock Anti-adversarially manipulated attributions for weakly and semi-supervised semantic segmentation.
\newblock In \emph{IEEE CVPR}, pages 4071--4080, 2021{\natexlab{b}}.

\bibitem[Lee et~al.(2022{\natexlab{a}})Lee, Oh, Yun, Choe, Kim, and Yoon]{lee2022weakly}
Jungbeom Lee, Seong~Joon Oh, Sangdoo Yun, Junsuk Choe, Eunji Kim, and Sungroh Yoon.
\newblock Weakly supervised semantic segmentation using out-of-distribution data.
\newblock In \emph{IEEE CVPR}, pages 16897--16906, 2022{\natexlab{a}}.

\bibitem[Lee et~al.(2022{\natexlab{b}})Lee, Kim, and Shim]{lee2022threshold}
Minhyun Lee, Dongseob Kim, and Hyunjung Shim.
\newblock Threshold {M}atters in {WSSS}: Manipulating the activation for the robust and accurate segmentation model against thresholds.
\newblock In \emph{IEEE CVPR}, pages 4330--4339, 2022{\natexlab{b}}.

\bibitem[Li et~al.(2022)Li, Fan, and Zhang]{li2022towards}
Jing Li, Junsong Fan, and Zhaoxiang Zhang.
\newblock Towards noiseless object contours for weakly supervised semantic segmentation.
\newblock In \emph{IEEE CVPR}, pages 16856--16865, 2022.

\bibitem[Liang et~al.(2022)Liang, Wang, Zhang, Sun, and Shen]{liang2022tree}
Zhiyuan Liang, Tiancai Wang, Xiangyu Zhang, Jian Sun, and Jianbing Shen.
\newblock Tree {E}nergy {L}oss: Towards sparsely annotated semantic segmentation.
\newblock In \emph{IEEE CVPR}, pages 16907--16916, 2022.

\bibitem[Lin et~al.(2014)Lin, Maire, Belongie, Hays, Perona, Ramanan, Doll{\'a}r, and Zitnick]{lin2014microsoft}
Tsung-Yi Lin, Michael Maire, Serge Belongie, James Hays, Pietro Perona, Deva Ramanan, Piotr Doll{\'a}r, and C~Lawrence Zitnick.
\newblock Microsoft {COCO}: {C}ommon objects in context.
\newblock In \emph{ECCV}, pages 740--755. Springer, 2014.

\bibitem[Liu et~al.(2022)Liu, Liu, Zhu, Shen, and Fernandez-Granda]{liu2022adaptive}
Sheng Liu, Kangning Liu, Weicheng Zhu, Yiqiu Shen, and Carlos Fernandez-Granda.
\newblock Adaptive early-learning correction for segmentation from noisy annotations.
\newblock In \emph{IEEE CVPR}, pages 2606--2616, 2022.

\bibitem[Oh et~al.(2021)Oh, Kim, and Ham]{oh2021background}
Youngmin Oh, Beomjun Kim, and Bumsub Ham.
\newblock Background-aware pooling and noise-aware loss for weakly-supervised semantic segmentation.
\newblock In \emph{IEEE CVPR}, pages 6913--6922, 2021.

\bibitem[Olsson et~al.(2021)Olsson, Tranheden, Pinto, and Svensson]{olsson2021classmix}
Viktor Olsson, Wilhelm Tranheden, Juliano Pinto, and Lennart Svensson.
\newblock Class{M}ix: Segmentation-based data augmentation for semi-supervised learning.
\newblock In \emph{IEEE WACV}, pages 1369--1378, 2021.

\bibitem[Pan et~al.(2021)Pan, Gao, Lin, Tang, Dong, Yuan, Huang, and Xu]{pan2021unveiling}
Xingjia Pan, Yingguo Gao, Zhiwen Lin, Fan Tang, Weiming Dong, Haolei Yuan, Feiyue Huang, and Changsheng Xu.
\newblock {U}nveiling the potential of structure preserving for weakly supervised object localization.
\newblock In \emph{IEEE CVPR}, pages 11642--11651, 2021.

\bibitem[Papandreou et~al.(2015)Papandreou, Chen, Murphy, and Yuille]{papandreou2015weakly}
George Papandreou, Liang-Chieh Chen, Kevin~P Murphy, and Alan~L Yuille.
\newblock {W}eakly-and semi-supervised learning of a deep convolutional network for semantic image segmentation.
\newblock In \emph{IEEE ICCV}, pages 1742--1750, 2015.

\bibitem[Park et~al.(2022)Park, Yang, Shin, Hwang, and Yang]{park2022saliency}
Joonhyung Park, June~Yong Yang, Jinwoo Shin, Sung~Ju Hwang, and Eunho Yang.
\newblock Saliency {G}rafting: Innocuous attribution-guided mixup with calibrated label mixing.
\newblock In \emph{AAAI}, pages 7957--7965, 2022.

\bibitem[Pinheiro and Collobert(2015)]{pinheiro2015image}
Pedro~O Pinheiro and Ronan Collobert.
\newblock {F}rom image-level to pixel-level labeling with convolutional networks.
\newblock In \emph{IEEE CVPR}, pages 1713--1721, 2015.

\bibitem[Qin et~al.(2022)Qin, Wu, Xiao, Li, and Wang]{qin2022activation}
Jie Qin, Jie Wu, Xuefeng Xiao, Lujun Li, and Xingang Wang.
\newblock Activation modulation and recalibration scheme for weakly supervised semantic segmentation.
\newblock In \emph{AAAI}, pages 2117--2125, 2022.

\bibitem[Rong et~al.(2023)Rong, Tu, Wang, and Li]{rong2023boundary}
Shenghai Rong, Bohai Tu, Zilei Wang, and Junjie Li.
\newblock Boundary-enhanced co-training for weakly supervised semantic segmentation.
\newblock In \emph{IEEE CVPR}, pages 19574--19584, 2023.

\bibitem[Rosenfeld and Pfaltz(1966)]{rosenfeld1966sequential}
Azriel Rosenfeld and John~L Pfaltz.
\newblock Sequential operations in digital picture processing.
\newblock \emph{Journal of the ACM (JACM)}, 13\penalty0 (4):\penalty0 471--494, 1966.

\bibitem[Roy and Todorovic(2017)]{roy2017combining}
Anirban Roy and Sinisa Todorovic.
\newblock {C}ombining bottom-up, top-down, and smoothness cues for weakly supervised image segmentation.
\newblock In \emph{IEEE CVPR}, pages 3529--3538, 2017.

\bibitem[Ru et~al.(2022)Ru, Zhan, Yu, and Du]{ru2022learning}
Lixiang Ru, Yibing Zhan, Baosheng Yu, and Bo Du.
\newblock Learning affinity from attention: end-to-end weakly-supervised semantic segmentation with transformers.
\newblock In \emph{IEEE CVPR}, pages 16846--16855, 2022.

\bibitem[Ru et~al.(2023)Ru, Zheng, Zhan, and Du]{ru2023token}
Lixiang Ru, Heliang Zheng, Yibing Zhan, and Bo Du.
\newblock Token contrast for weakly-supervised semantic segmentation.
\newblock In \emph{Proceedings of the IEEE/CVF Conference on Computer Vision and Pattern Recognition}, pages 3093--3102, 2023.

\bibitem[Shimoda and Yanai(2019)]{shimoda2019self}
Wataru Shimoda and Keiji Yanai.
\newblock {S}elf-supervised difference detection for weakly-supervised semantic segmentation.
\newblock In \emph{IEEE ICCV}, pages 5208--5217, 2019.

\bibitem[Su et~al.(2021)Su, Sun, Lin, and Wu]{su2021context}
Yukun Su, Ruizhou Sun, Guosheng Lin, and Qingyao Wu.
\newblock Context decoupling augmentation for weakly supervised semantic segmentation.
\newblock In \emph{IEEE ICCV}, pages 7004--7014, 2021.

\bibitem[Sun et~al.(2020)Sun, Wang, Dai, and Van~Gool]{sun2020mining}
Guolei Sun, Wenguan Wang, Jifeng Dai, and Luc Van~Gool.
\newblock Mining cross-image semantics for weakly supervised semantic segmentation.
\newblock In \emph{ECCV}, pages 347--365. Springer, 2020.

\bibitem[Wang et~al.(2020)Wang, Zhang, Kan, Shan, and Chen]{wang2020self}
Yude Wang, Jie Zhang, Meina Kan, Shiguang Shan, and Xilin Chen.
\newblock Self-supervised equivariant attention mechanism for weakly supervised semantic segmentation.
\newblock In \emph{IEEE CVPR}, pages 12275--12284, 2020.

\bibitem[Wu et~al.(2021)Wu, Huang, Gao, Wei, Wei, Luo, and Liu]{wu2021embedded}
Tong Wu, Junshi Huang, Guangyu Gao, Xiaoming Wei, Xiaolin Wei, Xuan Luo, and Chi~Harold Liu.
\newblock {E}mbedded discriminative attention mechanism for weakly supervised semantic segmentation.
\newblock In \emph{IEEE CVPR}, pages 16765--16774, 2021.

\bibitem[Xie et~al.(2022{\natexlab{a}})Xie, Hou, Ye, and Shen]{xie2022clims}
Jinheng Xie, Xianxu Hou, Kai Ye, and Linlin Shen.
\newblock {CLIMS}: Cross language image matching for weakly supervised semantic segmentation.
\newblock In \emph{IEEE CVPR}, pages 4483--4492, 2022{\natexlab{a}}.

\bibitem[Xie et~al.(2022{\natexlab{b}})Xie, Xiang, Chen, Hou, Zhao, and Shen]{xie2022c2am}
Jinheng Xie, Jianfeng Xiang, Junliang Chen, Xianxu Hou, Xiaodong Zhao, and Linlin Shen.
\newblock C2{AM}: Contrastive learning of class-agnostic activation map for weakly supervised object localization and semantic segmentation.
\newblock In \emph{IEEE CVPR}, pages 989--998, 2022{\natexlab{b}}.

\bibitem[Xie and Tu(2015)]{xie2015holistically}
Saining Xie and Zhuowen Tu.
\newblock Holistically-nested edge detection.
\newblock In \emph{IEEE ICCV}, pages 1395--1403, 2015.

\bibitem[Xu et~al.(2022)Xu, Ouyang, Bennamoun, Boussaid, and Xu]{xu2022multi}
Lian Xu, Wanli Ouyang, Mohammed Bennamoun, Farid Boussaid, and Dan Xu.
\newblock Multi-class token transformer for weakly supervised semantic segmentation.
\newblock In \emph{IEEE CVPR}, pages 4310--4319, 2022.

\bibitem[Yao et~al.(2021)Yao, Chen, Xie, Zhang, Shen, Wu, Tang, and Zhang]{yao2021non}
Yazhou Yao, Tao Chen, Guo-Sen Xie, Chuanyi Zhang, Fumin Shen, Qi Wu, Zhenmin Tang, and Jian Zhang.
\newblock {N}on-salient region object mining for weakly supervised semantic segmentation.
\newblock In \emph{IEEE CVPR}, pages 2623--2632, 2021.

\bibitem[Yun et~al.(2019)Yun, Han, Oh, Chun, Choe, and Yoo]{yun2019cutmix}
Sangdoo Yun, Dongyoon Han, Seong~Joon Oh, Sanghyuk Chun, Junsuk Choe, and Youngjoon Yoo.
\newblock {CutMix}: {R}egularization strategy to train strong classifiers with localizable features.
\newblock In \emph{IEEE ICCV}, pages 6023--6032, 2019.

\bibitem[Zhang et~al.(2020{\natexlab{a}})Zhang, Xiao, Wei, Sun, and Huang]{zhang2020reliability}
Bingfeng Zhang, Jimin Xiao, Yunchao Wei, Mingjie Sun, and Kaizhu Huang.
\newblock Reliability {D}oes {M}atter: {A}n end-to-end weakly supervised semantic segmentation approach.
\newblock In \emph{AAAI}, pages 12765--12772, 2020{\natexlab{a}}.

\bibitem[Zhang et~al.(2020{\natexlab{b}})Zhang, Zhang, Tang, Hua, and Sun]{zhang2020causal}
Dong Zhang, Hanwang Zhang, Jinhui Tang, Xian-Sheng Hua, and Qianru Sun.
\newblock Causal intervention for weakly-supervised semantic segmentation.
\newblock \emph{NeurlPS}, 33:\penalty0 655--666, 2020{\natexlab{b}}.

\bibitem[Zhang et~al.(2021)Zhang, Gu, Zhang, and Dai]{zhang2021complementary}
Fei Zhang, Chaochen Gu, Chenyue Zhang, and Yuchao Dai.
\newblock Complementary patch for weakly supervised semantic segmentation.
\newblock In \emph{IEEE ICCV}, pages 7242--7251, 2021.

\bibitem[Zhou et~al.(2016)Zhou, Khosla, Lapedriza, Oliva, and Torralba]{zhou2016learning}
Bolei Zhou, Aditya Khosla, Agata Lapedriza, Aude Oliva, and Antonio Torralba.
\newblock Learning deep features for discriminative localization.
\newblock In \emph{IEEE CVPR}, pages 2921--2929, 2016.

\bibitem[Zhou et~al.(2022)Zhou, Zhang, Zhao, and Li]{zhou2022regional}
Tianfei Zhou, Meijie Zhang, Fang Zhao, and Jianwu Li.
\newblock Regional semantic contrast and aggregation for weakly supervised semantic segmentation.
\newblock In \emph{IEEE CVPR}, pages 4299--4309, 2022.

\end{thebibliography}
}


\end{document}